\documentclass[preprint,12pt]{elsarticle}

\usepackage{ragged2e}
\usepackage{amsmath, amssymb}
\usepackage{mathrsfs}
\usepackage{algorithm}
\usepackage{algpseudocode}
 \usepackage{graphicx}
 \usepackage{tabularx,booktabs}
 \usepackage{amsfonts}
\usepackage{array}
\usepackage{supertabular,booktabs}
\usepackage{float}
\usepackage{longtable}
\usepackage{lipsum}
\usepackage{xcolor}
\usepackage{multicol}
\usepackage{multirow}
\usepackage{url}
\usepackage{caption}
\usepackage{ltablex}
\usepackage{bm}

\journal{}

\begin{document}

\begin{frontmatter}



\title{Decision-Making-Based Path Planning for Autonomous UAVs: A Survey}



\author[1]{*Kelen C. Teixeira Vivaldini\corref{cor1}}



\ead{vivaldini@ufscar.br}

\author[2]{Robert Pěnička}
\ead{penicrob@fel.cvut.cz}

\author[2]{Martin Saska}
\ead{martin.saska@fel.cvut.cz}

\affiliation[2]{organization={Department of Cybernetics, Faculty of Electrical Engineering, Czech Technical University in Prague},
    addressline={Karlovo Namesti 13, Prague 2}, 
    city={Prague},
    postcode={CZ 121 35}, 
    country={Czech Republic}}

\affiliation[1]{organization={Department of Computer, Federal University of São Carlos},
    addressline={Rod. Washington Luís, Km 235}, 
    city={São Carlos},
    postcode={13565-905}, 
    state={São Paulo},
    country={Brazil}}

\cortext[cor1]{Corresponding author}

\begin{abstract}

One of the most critical features for the successful operation of autonomous UAVs is the ability to make decisions based on the information acquired from their surroundings. 
Each UAV must be able to make decisions during the flight in order to deal with uncertainties in its system and the environment, and to further act upon the information being received. Such decisions influence the future behavior of the UAV, which is expressed as the path plan. 
Thus, decision-making in path planning is an enabling technique for deploying autonomous UAVs in real-world applications. This survey provides an overview of existing studies that use aspects of decision-making in path planning, presenting the research strands for Exploration Path Planning and Informative Path Planning, and focusing on characteristics of how data have been modeled and understood. Finally, we highlight the existing challenges for relevant topics in this field.
\end{abstract}

\begin{graphicalabstract}

\end{graphicalabstract}

\begin{highlights}
    \item Survey of existing studies that use aspects of decision-making in path planning for autonomous UAVs.
    \item Analysis of situations with limited information leading to uncertainty in decision-making.
    \item Overview of works solving these topics and identification of research gaps.
    \item Exploration and Informative Path Planning strategies analyzed and reviewed.
    \item Discussion of challenges of dealing with uncertainty and potential solutions.


\end{highlights}


\end{frontmatter}



\section{INTRODUCTION}
\label{sec:introduction}

Effective path planning is critical for the autonomous operation of UAVs (Unmanned Aerial Vehicles), enabling them to navigate through complex, dynamic, and uncertain environments to perform tasks without human intervention. A key aspect of this process is the decision-making, which allows the UAV to select actions that optimize mission objectives while adapting to new information and environmental changes~\cite{zhao2018brain, AGGARWAL2020270}.

In missions where the environment is only partially known or constantly changing, decision-making-based path planning emerges as a necessary approach. In this paradigm, path planning becomes a sequence of interdependent decisions, rather than a one-time computation, where the UAV must dynamically answer: \emph{"Where should I go next?"}, \emph{"Which areas should be prioritized for exploration or data collection?"}, and \emph{"When should the mission be terminated or redirected?"}.
These decisions must consider factors such as information gain, mission objectives, operational constraints, environmental uncertainty, risk levels, and time or energy limitations. Under uncertainty, path planning transforms into a decision-making problem, where each action affects future possibilities and outcomes.

Traditional path planning methods typically compute a static, optimal route between predefined points\cite{Dijkstra59}, often neglecting the dynamic value of new information gathered during the mission. In contrast, decision-making-based planning adapts the path, allowing the UAV to prioritize areas of higher relevance or uncertainty in real-time.

\begin{figure}[H]
    \centering
   
    \includegraphics[width=0.47\textwidth]{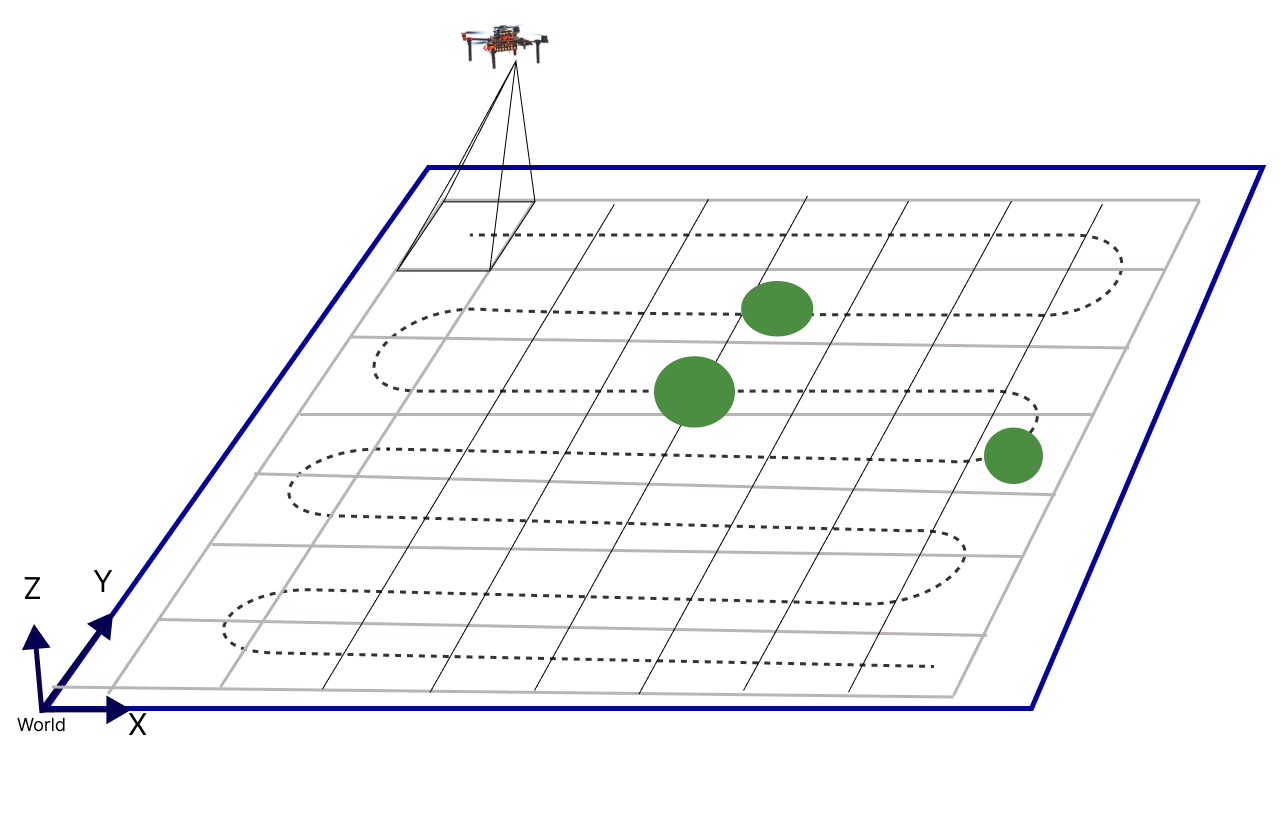}
    \includegraphics[width=0.47\textwidth]{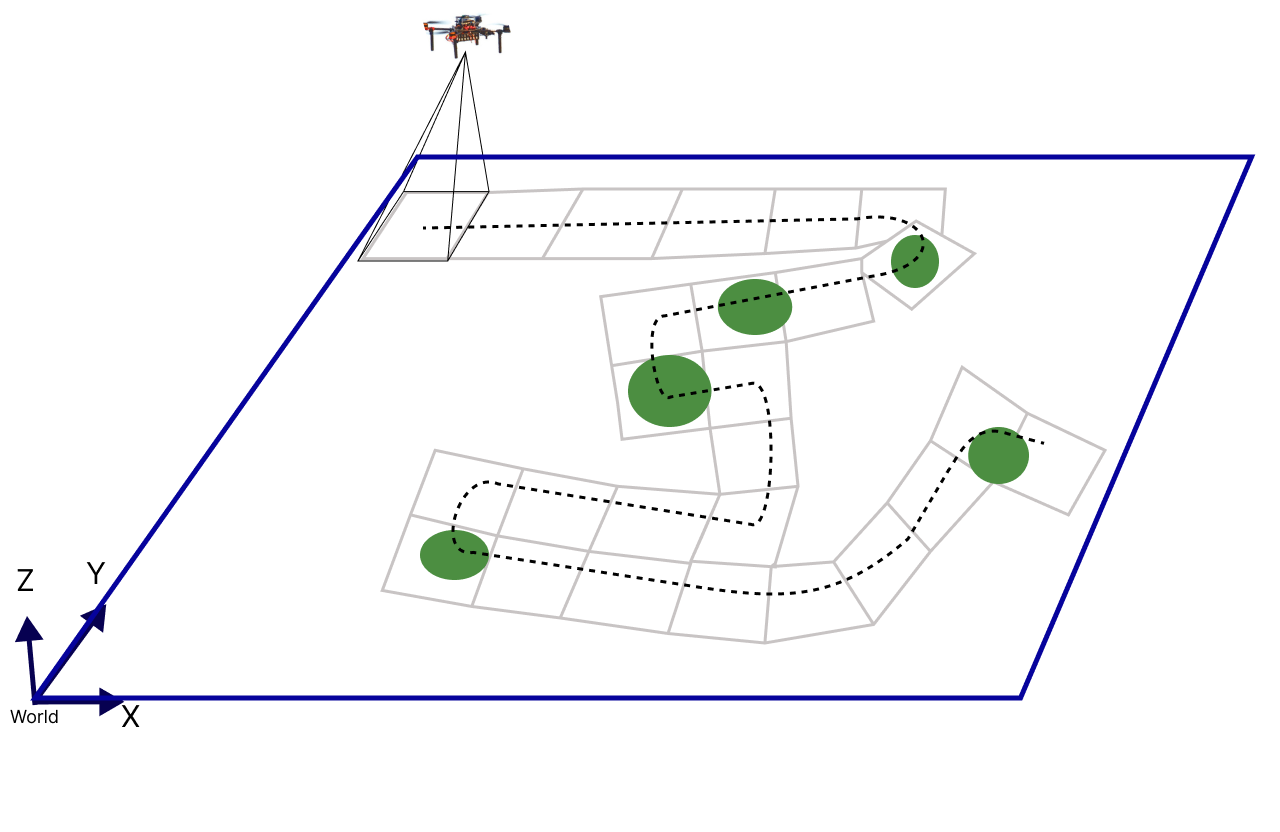}
     \\
      (a)\qquad\qquad\qquad\qquad\qquad\qquad(b)
     \caption{Comparison of path planning strategies. (a) Predefined path planning: the UAV follows a fixed path regardless of real-time feedback, and (b) Decision-making-based path planning: the UAV adapts its path based on acquired data to improve mission performance.}
    \label{fig:diff}
\end{figure}

Figure~\ref{fig:diff} illustrates this distinction. In conventional path planning, Fig. \ref{fig:diff} (a)), the UAV follows a predetermined path, regardless of the data collected covering an area and/or avoiding obstacles, and does not leverage real-time data to reoptimize its route or maximize information gain. Conversely, in decision-making-based planning (Fig. \ref{fig:diff} (b)), the UAV continuously adapts its path based on observations, prioritizing high-value areas and avoiding uninformative regions.

This adaptive behavior is funcamental in applications such as environmental monitoring, where UAVs select sensing locations to characterize better a phenomenon/target~\cite{Souza:2015,Gonzalez2017,Popovic2017online,Popovic2020_IPP,li2022multi}; surveillance and search and rescue, where the UAV must rapidly decide where to search next under uncertainty~\cite{Peng2018,Meng2017,Li2022,Boulares2021,Yifei2022}; and air quality and temperature mapping, where the UAV must measurement in spatially and temporally variable conditions~\cite{Xia2019drone,Shen2020SynergisticPP,Jin2022,Ruckin2022}. 

These missions require UAVs to make decisions that balance exploration (gathering new knowledge) and exploitation (refining existing knowledge), often under tight constraints.  
This makes decision-based path planning a computationally and conceptually challenging problem, due to: (i) the dynamic and uncertain nature of the environment; (ii) sensor and communication limitations; (iii) planning under uncertainty requires modeling, predicting, and optimizing over a high-dimensional decision space in real time; and (iv) conflicting objectives, such as maximizing information gain while minimizing energy consumption or risk exposure.

Notably, many works implement decision-making principles even when not explicitly labeled as such. In this context, two paradigms have gained prominence: \textit{Exploration Path Planning} (EPP) and \textit{Informative Path Planning} (IPP). Both rely on decision-making principles to guide UAVs, yet differ in emphasis and modeling approach.
EPP focuses on incrementally building an accurate map of the environment while planning to revisit or expand into regions that minimize overall representation uncertainty \cite{stachniss2009robotic,francis2020functional,francis2019occupancy}. Key decision metrics include path cost, information gain, and fulfillment of application requirements~\cite{CHOUDHARY2018}.

IPP seeks paths that maximize expected information gain subject to budget constraints~\cite{Moon2022,Popovic2020_IPP}. IPP addresses the exploration–exploitation trade‑off by leveraging current beliefs to guide data collection efficiently \cite{McCammon2021}, often using uncertainty‑reduction techniques ~\cite{marchant2014bayesian}.

Before delving into each paradigm, it is necessary to present the foundational concepts of mapping and monitoring, as these tasks underpin most decision-making-based planning strategies in autonomous UAV missions. 

\emph{Mapping} refers to exploring an environment while learning its spatial structure. Many existing approaches emphasize spatial coverage and consistency, sometimes neglecting the quality or completeness of the map representation. When focusing solely on coverage, the decision to terminate exploration is straightforward; however, ensuring high-quality representations increases complexity due to the large required data volume~\cite{lluvia2021}. In this context, active mapping employs UAVs with onboard sensors to autonomously scan and reconstruct continuous 2D or 3D environments~\cite{Popovic2020_IPP}. Decision-making-based path planning enhances this process by allowing the UAV to continuously evaluate and select actions that optimize the path and the fidelity of the resulting map, adapting to new data and environmental uncertainty. In this sense, mapping is not just about where to fly but about making informed decisions during flight to maximize map quality and mission utility.

\emph{Monitoring}, on the other hand, focuses on observing an environment over time to quantify ongoing changes~\cite{Cambridge}. UAVs are particularly effective in this context because they can perform spatiotemporal sampling across large or inaccessible areas~\cite{Dunbabin2012}. Monitoring strategies range from discrete sampling at predefined locations, which may limit coverage efficiency~\cite{Gao2022}, to continuous data acquisition along the UAV's path. In either case, decision-making-based path planning enables the UAV to adjust its behavior based on observed environmental dynamics, optimizing the relevance and quantity of the collected data.

In this context, the boundaries between \emph{Exploration Path Planning} (EPP) and \emph{Informative Path Planning} (IPP) often become indistinct, as both paradigms rely on the UAV's ability to make informed decisions that maximize utility under uncertainty. According to the literature, a substantial conceptual overlap between these approaches, particularly when UAVs autonomously select trajectories to optimize information gain. This article contributes to understanding such strategies by systematically reviewing decision-making-based planning methods and highlighting their application in autonomous UAV operations. We clarify the distinctions and commonalities between EPP and IPP, discuss their connection to mapping and monitoring tasks, and summarize the primary methods developed to address these challenges.

In summary, this paper contributes to the scientific understanding of decision-making-based path planning for autonomous UAVs by:   

1) Providing a systematic review of the state-of-the-art decision-making-based path planning, focusing on the EPP and IPP paradigms. 

2) Analyzing how uncertainty, due to limited or missing prior information, impacts decision outcomes and planning strategies.

3) Identifying research gaps and open challenges, offering insights for future developments in adaptive and autonomous UAV navigation.

\section{Path Planning and Autonomous UAVs}
\label{sec:PathPlanning_UAV}

Regardless of the mission, path planning is a fundamental component of autonomous UAV operation. Its primary goal is to guide the system from an initial, suboptimal state to a more desirable one~\cite{Lavalle2000}. Classical algorithms, including sampling-based or optimization-based methods, aim to compute feasible and efficient paths while considering constraints such as obstacle avoidance and kinematics. Benchmarks such as PathBench~\cite{toma2021pathbench} and Planie~\cite{RochaPlannie,RochaBenchmark} are commonly used to evaluate these algorithms based on performance indicators like robustness, completeness, computational cost, memory usage, and battery consumption. Widely adopted platforms such as OMPL (Open Motion Planning Library)~\cite{sucan2012open}, MoveIt~\cite{moll2015benchmarking}, and OOPS (Online, Open-source Programming System)~\cite{plaku2007oops} also incorporate these metrics to benchmark motion planning strategies.

Although classical algorithms allow the selection of efficient strategies for specific applications, they are not sufficient on their own to enable full autonomy. Autonomous behavior requires additional mechanisms to guide the UAV based on the expected benefit of acquiring new information from the environment. In this context, the concept of \textit{information gain} plays a central role. It refers to the amount of information the UAV expects to obtain by moving from its current position to a new one, and it is often related to the exploration of unknown areas. To achieve efficient path planning, the decision-making process must therefore maximize information gain while also accounting for movement costs, such as distance, time, or energy consumption~\cite{petit2022tape,Stachniss2005,bai2016information,tabib2016computationally}.
As a result, research areas such as \textit{Informative Path Planning} and \textit{Exploration Path Planning} have emerged, which integrate decision-making-based methods with traditional path planning techniques. These approaches allow UAVs to choose not only feasible paths but also those that are most advantageous for acquiring valuable information, leading to more efficient and intelligent autonomous missions.

\section{Informative Path Planning} 
\label{sec:Robotic_IPPP}

\textit{Informative Path Planning} (IPP) refers to a class of decision-making path planning problems where the objective is not merely to reach a destination but to autonomously select paths that maximize the information gain, subject to resource constraints such as time, energy, or mission-specific objectives~\cite{Orisatoki2025informative,Popovic2020_IPP,Meliou_NIPP2007,Arora2017}. Unlike classical path planning, e.g., Dijkstra’s algorithm~\cite{Dijkstra59}, which aims to find the shortest or least-cost path, IPP prioritizes informational efficiency, guiding the robot toward paths that reduce uncertainty through sequential decisions~\cite{Orisatoki2025informative}.

A defining feature of IPP is its sequential and adaptive decision-making process. As a UAV collects sensor data, it updates its internal model of the environment and adapts its path accordingly. This forms a dynamic feedback loop, distinguishing IPP from static planning approaches. Autonomous decisions are made based on previous observations, which directly influence the effectiveness of future measurements. This adaptivity is central to maximizing information gain under uncertainty~\cite{aniceto2022}.

To support this adaptive planning, IPP frameworks typically employ on probabilistic models, such as occupancy grids, Gaussian processes, or belief maps, to represent and reason about environmental uncertainty. These models enable the computation of an objective function that estimates the expected information gain from future actions, guiding toward the most informative regions of the environment.


This approach contrasts with Coverage Path Planning (CPP), which aims at complete area coverage. While CPP is suitable for exhaustive surveying tasks, IPP focuses on targeted exploration, selectively prioritizing areas that offer higher expected information gain. As such, IPP is particularly advantageous in resource-constrained missions where only limited sensing actions are possible, such as disaster response, search and rescue, environmental monitoring, or hazardous area inspection.

Furthermore, in these active sensing scenarios, the number of measurements that can be performed is constrained by the mission's resource budget. Consequently, the total amount of acquirable information is limited. In this context, the utility of new observations typically follows submodularity behavior, where the marginal gain from acquiring additional data decreases as more information is collected, due to spatial and temporal correlations in the environment~\cite{krause2011submodularity}. This property reflects real-world sensing dynamics and offers opportunities for algorithmic approximations in IPP problem formulations.

Such submodularity leads to a non-linear objective space, making the optimization problem inherently complex, especially in large or unstructured environments~\cite{Popovic2020_IPP}. This way, the IPP problem is generally classified as NP-Hard \cite{singh2009efficient}\cite{Arora2017}\cite{Hollinger2014}, as the size of the space of the information collected along the path increases exponentially during mission execution. And in some cases, even PSPACE-hard\cite{reif1979complexity}, depending on the form of the utility function and the path representation.

The reward function used in IPP is often non-convex, containing multiple local optima \cite{McCammon2021,Popovic2017multi}, and depends heavily on prior knowledge and environment structure. In the literature, IPP is often associated with information-gathering tasks in cost-constrained or continuous action spaces~\cite{Arora2017,Gan2014,Hollinger2014}. The core computational challenge lies in evaluating the expected information gain of candidate paths and selecting the most informative path \cite{Marchant2014}.

Formally, IPP can be defined as the problem of finding an optimal path within the space of all possible continuous paths that maximizes an information gain, subject to a budget constraint~\citet{Popovic2020_IPP}. Thus, IPP enables UAVs to autonomously plan paths based on sensory observations by continuously evaluating the expected utility of future measurements and selecting paths that maximize information gain~\cite{Marchant2014}. This leads to the development of both \textit{myopic} and \textit{non-myopic} strategies.

In the myopic approach, the UAV decides on its following movements sequentially, optimizing the immediate reward at each step~\cite{Blanchard2022,Stachniss2005,Marchant2014}. These strategies typically rely on greedy heuristics, evaluating actions based solely on short-term gains~\cite{Marchant2012,Charrow2015,Viseras2019}. For instance, the best action is chosen based on which candidate location is expected to offer the highest immediate information gain. While computationally efficient and suitable for time-constrained scenarios, myopic planners often yield suboptimal global solutions because they fail to anticipate long-term consequences~\cite{Morere2017}. 

An example is the recursive greedy algorithm for a graph-based IPP formulation proposed by~\citet{Binney2013}, which incorporates a time-varying objective function and evaluates edge-wise observations to guide exploration over a graph structure.
Although sufficient in many applications, myopic methods become inadequate when current decisions can negatively impact long-term performance, precisely due to their lack of anticipation mechanisms~\cite{Blanchard2022}.

To address such limitations, \citet{Blanchard2022} proposed a Bayesian Optimization (BO) framework for anomaly detection in exploration missions. Their method introduces a probabilistic model that eliminates ad-hoc parameters and incorporates a Bayesian update mechanism. This update allows the UAV to learn and adapt its behavior based on the environment, improving planning through an acquisition function that balances exploration and exploitation. Comparative analysis revealed that likelihood-weighted acquisition functions yielded superior performance in environments with incomplete or incorrect prior information.

Improving upon purely myopic behavior, \citet{Nguyen2016} employed a finite-horizon tree search to enhance planning performance, even with minimal prior knowledge. They proposed the Cluster Tree Search (CTS) method, which constructs high-utility plans by combining local subplans formed around information clusters. CTS has two variants: CTS-Seq (Sequential composition) and CTS-DP (Dynamic Programming-based composition). Both achieved high-quality planning with reduced computational cost, and CTS-DP was validated through real-time experiments.
The primary difference between myopic and non-myopic planning lies in foresight. Myopic planners evaluate single-step rewards, which makes them vulnerable to local optima. 

In contrast, non-myopic strategies consider sequences of decisions and aim to optimize over a planning horizon~\cite{Blanchard2022,Meliou_NIPP2007,Morere2017,singh2009efficient,morere2016bayesian,Meera2019}. This enables them to escape local minima and generate more globally optimal plans. It considers the entire sequence of decisions and evaluates the overall cost associated with the sequence, instead of considering only the next destination. The non-myopic strategy is usually more efficient than myopic strategies since it takes into account the overall cost of all the destinations in the sequence, as opposed to just the next destination. 

Non-myopic planning utilizes a broader range of methods, including graph search~\cite{Binney2012,Zhu2021,Ma2018,Kodgule2019}, random trees~\cite{Viseras2019,Yang2013,Schmid2020,Xiong2020}, evolutionary algorithms~\cite{Popovic2020_IPP,Hitz2017,Xiong2019}, finite-horizon planning~\cite{Cao2013,Lim2016adaptive,Batinovic2022shadowcasting,Fernandez2022}, adaptive strategies~\cite{Choudhury2020}, and mixed-integer programming~\cite{Dutta2022}. A typical modeling method is the Partially Observable Markov Decision Process (POMDP), which enables decision-making under uncertainty using beliefs about the hidden state and rewards for sequences of actions.

Non-myopic strategies traditionally compute complete path sequences offline, aiming to generate optimal policies by anticipating the most likely future scenarios. While theoretically promising, this approach often degrades in performance due to reliance on potentially inaccurate models or when little prior data is available. The core limitation lies in the inability to refine decisions based on newly acquired information during execution. In contrast, online non-myopic strategies, such as the one presented by~\citet{Fernandez2022}, recalculate the optimal action after each path using simulated sequences informed by real-time data. This allows UAVs to sample from a potentially infinite set of future trajectories. In dynamic, stochastic, or noisy environments, this makes the problem of selecting the best action a non-trivial maximization task~\cite{Morere2017}.

Recent studies have demonstrated that combining IPP strategies with semantic scene understanding improves performance in real-world operations. As discussed in~\cite{Orisatoki2025informative}, contextual information about the environment—e.g., terrain type, object classes, or visibility conditions—can be incorporated into the reward function to guide exploration in unstructured or dynamic environments better. The integration of such semantic data into the planning pipeline allows UAVs to prioritize data collection in areas of higher interest or uncertainty, which is particularly relevant in missions where the quality of gathered information varies spatially. Furthermore, ~\cite{Orisatoki2025informative} emphasizes that in long-horizon missions, adaptive strategies that fuse real-time feedback with prior semantic maps achieve more consistent performance than purely model-based approaches. This reinforces the need for flexible planning methods that can handle changing operational contexts. 

Submodularity—a property that arises when spatial or temporal correlations exist among measurements—implies that past observations can inform current ones. As noted by~\citet{Popovic2020informative}, this property makes the objective space highly nonlinear and the resulting optimization problem combinatorial, which significantly complicates planning in large and complex environments.

Several works have proposed IPP as a sequential decision-making problem. Algorithms such as Greedy, Recursive Greedy~\cite{singh2009efficient}, and Branch-and-Bound~\cite{Binney2012} have been applied. However, they often evaluate actions in isolation, failing to account for the cumulative impact of sequences of actions—a limitation especially relevant in non-myopic planning. Sampling-based methods like Rapidly-exploring Information Gathering (RIG)~\cite{Hollinger2014} partially address this by building trees over the global information field, though individual actions are still evaluated separately.

\citet{Arora2017} proposed the Randomized Anytime Orienteering-Greedy (RAOr-G) algorithm, which improves planning by restricting local search to admissible sets determined via the Travelling Salesman Problem (TSP), within a defined budget. Weighted sampling and local greedy heuristics increase the likelihood of visiting high-value nodes, improving runtime performance in scenarios with low probabilities of near-optimal paths, but at the expense of theoretical guarantees.

Other approaches enhance IPP by modifying the BO objective function or refining the Gaussian Process (GP) model to account for discrete or continuous input characteristics~\cite{Salhotra2021}. One example is a BO-based strategy developed for monitoring eucalyptus diseases~\cite{vivaldini2016}, where GP regression is used to generate a pixel-based map and guide UAVs toward uncertain regions with high classification potential, minimizing variance through a tailored acquisition function. This approach maximizes the search area and minimizes uncertainty about the diseased trees.

An active classification framework based on Covariance Matrix Adaptation Evolution Strategy (CMA-ES) has also been proposed~\cite{Popovic2017online}. The method uses smooth polynomial trajectories and a height-dependent sensor model to ensure feasibility under UAV dynamic constraints. Compared to RIG-tree and coverage planners, the CMA-ES approach reduces map entropy and improves classification rates. However, its scalability is limited due to the high computational cost of evolutionary optimization.

Building on this framework, an extension targeting multi-resolution mapping for agricultural monitoring was proposed by~\citet{Popovic2017multi}. The authors adopted a fixed-horizon algorithm to generate informative and dynamically feasible trajectories in continuous 2D or 3D space, using an informed initialization. The approach leverages spatial correlations modeled by Gaussian Processes and fuses data from dependent probabilistic sensors via Bayesian methods. Regular measurement updates allow consistent map refinement, resulting in a 45\% error reduction compared to the RIG-tree method in simulations.

The Obstacle-aware Adaptive Informative Path Planning algorithm for UAV target search addresses a layered optimization scheme~\cite{Meera2019}. Based on~\cite{Popovic2017multi} and incorporating adaptive BO planning from~\cite{Brochu2010}, it optimizes a smooth and continuous trajectory to maximize information gain while balancing exploration and exploitation, maintaining robustness against false detections.

A further extension of the BO-based approach focuses on active classification under uncertainty~\cite{vivaldini2019uav}. This strategy maximizes knowledge of visited areas and minimizes classification uncertainty of diseased trees, while incorporating a return-to-base constraint in the acquisition function to ensure optimal flight time utilization.

Resource-aware data collection strategies for UAVs in uncertain environments have been introduced using a BO-based method with the Time-weighted Expected Improvement acquisition function~\cite{Popovic2020informative}. This strategy maximizes information gain while accounting for resource constraints such as battery life, mission time, and path length.

Anomaly detection in environmental exploration has also benefited from BO-based IPP strategies~\cite{Blanchard2022}. The method integrates prior beliefs about regions of interest, refining them through GP regression during exploration. A Dubins path ensures continuous paths respecting turning constraints. The next-best view is selected by a modified BO function, with a gain policy (proposed by~\cite{Morere2017}) that guides the UAV toward high-uncertainty regions.

A POMDP-based approach to IPP has been developed to focus on estimating quantiles rather than ensuring uniform coverage~\cite{Fernandez2022}. By adapting the Partially Observable Monte Carlo Planner (POMCP) planner~\cite{Silver2010}, the authors introduce novel objective functions—quantile change and quantile standard error—that outperform traditional information-theoretic and entropy-based baselines. The proposed objective function aims to minimize model reconstruction error while simultaneously maximizing environmental coverage.

Adaptive sampling has been applied to IPP to efficiently plan paths in high-dimensional and large search spaces~\cite{Moon2022}. This method constructs a tree from sampled points within the search space to guide the path planning, aiming to maximize information gain and reduce entropy over the belief space under a limited budget. However, as the search space grows, performance may degrade because maintaining sufficiently dense trees becomes computationally intractable.

Reinforcement Learning (RL) strategies with adaptive online replanning capabilities have been proposed to address IPP in 3D environments~\cite{Ruckin2022}. Unlike previous RL-based works~\cite{Viseras2019, chen2020autonomous} that focus on limited 2D areas, this method integrates Monte Carlo Tree Search (MCTS) with Convolutional Neural Networks (CNNs) to explore and map regions of higher information value in complex, unknown environments.

Table~\ref{tab:works_IPP} summarizes the main contributions in the state-of-the-art for solving the IPP problem.

\begin{center}
{\footnotesize
\begin{longtable}[H]{p{3cm}p{4cm}p{5.2cm}}

\caption{Techniques solving the IPP.} \label{tab:works_IPP} \\

\hline \textbf{Author}& {\textbf{Methods}} & {\textbf{Objective}} \\ \hline 
\endfirsthead
\multicolumn{3}{l}{{Continuation - Techniques solving the IPP}} \\
\hline \textbf{Author}& {\textbf{Methods}} & {\textbf{Objective}}  \\ \hline 
\endhead

\hline \multicolumn{3}{l}{{Continued on next page}} \\ \hline
\endfoot

\hline \hline
\endlastfoot
\citet{Singh2007}& Recursive-greedy algorithm with Branch and Bound

Non-stationary Gaussian Process model& 
Proposed an efficient multi-robot IPP strategy using submodularity (mutual information) to coordinate robots under resource constraints for spatial data collection.\\ \hline

\citet{Cao2013}& Gaussian Process (GP)
  
MEEP (Maximum Entropy Path Planning)
  
Maximum Mutual Information Path Planning

&  
Investigated entropy-based vs. mutual-information-based path planning for active sensing; introduced MEPP to balance sensing performance and time.\\ \hline 

\citet{Hollinger2014} & Sampling-based methods

GP models

Continuous space RIG-roadmap

RIG-tree and RIG-graph & 
Developed motion-aware IPP in continuous domains using roadmap-guided exploration and Gaussian Processes.\\ \hline 

\citet{Marchant2014} & Bayesian Optimization 

GP  & 
Applied sequential Bayesian optimization to explore spatiotemporal phenomena and maximize information gain.
\\ \hline 

\citet{Gan2014}& Augmented Lagrangian

Monte Carlo 

Decentralized optimization& 
Designed an online decentralized optimization framework for constrained, real-time target search in dynamic environments.\\ \hline 

\citet{Bostrom2019}& Extended Kalman filter (EKF)

Greedy and graph-based search

Monte Carlo planning   & Proposed a deterministic sampling method for informative planning under target motion uncertainty. 
\\ \hline 

\citet{Cliff2015} & Grid-based filtering

Greedy IPP

Improved greedy IPP to refine confidence in predicted target positions while considering uncertainty within fixed observation ranges.\\ \hline 
   
\citet{Nguyen2016} & Adapt Monte Carlo Tree Search (MCTS)

GP

Cluster Tree Search (CTS) & 
Addressed energy-aware IPP for gliders, leveraging wind fields to sustain flight while collecting valuable information.
\\ \hline 

\citet{Lim2016adaptive} & Recursive Adaptive Identification (RAId)
   
Finite-horizon optimization &  
Introduced an adaptive trade-off framework for information gain versus path cost in metric spaces.\\ \hline 

\citet{vivaldini2016} & Discrete and Continuous Bayesian Optimization

Logistic Regression

Gaussian Process Regression (GPR) &  
Proposed a UAV-based IPP for maximizing classification coverage while minimizing target uncertainty in spatial data acquisition.\\ \hline 
\citet{Cho2018} &  Multiple Depots Multiple Traveling Salesmen Problem 

Gaussian Process Regression 

KL-divergence & An IPP considering the wind field dynamics to map a specific region using a set of UAVs. \\ \hline 

\citet{Arora2017} & Constraint Satisfaction Problem (CSP)
  
Asymptotically optimal IPP algorithm
  
Gaussian Process 
  
TSP & Near-optimal solutions to obtain the paths that are maximized regarding the subject of reward functions and constraints on path length. \\ \hline 

\citet{Popovic2017online} & 3D grid search 
   
CMA-ES (Covariance Matrix Adaptation - Evolution Strategy) & Information-rich trajectories in continuous 3D space with lower uncertainty.  \\ \hline 

\citet{Popovic2017multi} & 
CMA-ES 
    
& Dynamically feasible trajectories for maximum informativeness in continuous space, considering a budget and sensing constraints. \\ \hline 

\citet{Hitz2017} & Continuous-space Informative Path Planner (CIPP) and CMA-ES & A real-world adaptive path replanning to identify and explore regions of interest in the scalar field.\\ \hline 

\citet{Morere2017} & Partially Observable Markov Decision Process (POMDP)
    
Sequential Bayesian Optimization
    
Monte-Carlo Tree Search (MCTS) & A myopic planner based on the BO-POMDP algorithm, reformulating the reward function 

\cite{marchant2014bayesian} to balance the trade-off between exploitation and exploration of high-gradient areas.\\ \hline 

\citet{Meera2019} & Obstacle-aware Adaptive Informative Path Planner (OA-IPP) & Optimize a smooth continuous trajectory for maximal information gain. \\ \hline

\citet{vivaldini2019uav} & Continuous Bayesian Optimization
  
Gaussian Process Regression
  
Logistic Regression 
  
Rapidly exploring Random Tree (RRT) & A route planning Bayesian Optimization technique for active classification under uncertain conditions to maximize the information gain and minimize the uncertainties, considering the distance traveled constraint. \\ \hline 

\citet{Choudhury2020} & Adaptive IPP with Multimodal Sensing

POMDP - Monte Carlo 

Tailored adaptive greedy roll-out policy  &  Additional complexity of deciding between sensors with varying energy costs and observation models, appropriate structure to jointly reason
about movement, and sensing.\\ \hline 

\citet{Popovic2020informative} & CMA-ES 
   
   SegNet convolutional architecture  & A framework for mapping discrete or continuous variables and feasible trajectories in continuous space.\\ \hline

\citet{Velasco2020} & Adaptive Cascaded Local Optimal Planning & An adaptive multi-resolution IPP based on Gas Distribution Modelling (GDM) for high quality and efficient sampling in known and unknown environments. \\ \hline 

\citet{Brouwer2021} & B-splines 
   
   Gaussian Process &  A balance between exploration of the workspace and exploiting a source-seeking behavior to plan paths around known obstacles and detecting radioactive sources. \\   \hline

\citet{McCammon2021} & Hierarchical Hotspot Information Gathering (HHIG) and Topology-Aware Self-Organizing Map & A hierarchical informative path planner to incorporate the global information about the topological structure of the robot’s information space, and reduce the computation for effective solutions. \\ \hline 

\citet{Salhotra2021} & POMDPs

MCTS &  Improvements in roll-out allocation, the action exploration algorithm, and plan commitment.\\ \hline 
   
\citet{Zhu2021}& Manifold Gaussian processes 

Geodesic kernel functions
   
   Receding horizon planning & 3D coverage planning and random exploration to reconstruct error and information-theoretic metrics on three-dimensional surfaces.
\\ \hline 

\citet{Batinovic2022shadowcasting} &  Receding Horizon Next-Best-View (NBV) planning 

RRT 

Recursive Shadowcasting (RSC)  & A method designed to explore an area while planning paths and calculating information gain in a short computation time.\\ \hline 

\citet{Blanchard2022}   & Bayesian Optimization 
    
    Gaussian Process  &  Reconstruction of seafloor topography from real-world bathymetry data.\\ \hline 

\citet{Ercolani2022}& 3D Kernel DM+V/W algorithm 
   
   Kullback-Leibler Divergence Clustering & A clustering strategy to enhance exploration.\\ \hline 

\citet{Fernandez2022} &  Partially Observable Monte Carlo Planner (POMCP) 
   
   Gaussian Process & An adaptive planner to select locations for sample collection, aiming to maximize spatial coverage. \\ \hline

\citet{Ghassemi2022} & Bayes-Swarm-P Algorithm
   
   COBYLA/L-BFGS-B combination
   
   PSO (Particle Swarm Optimization) & A dynamic adaptation of the exploitation-exploration balance, improving search efficiency and convergence in a decentralized swarm.
   \\ \hline 
   
 \citet{Li2022} & DCEE
   
   Bayesian Optimization 
   
   Particle filter  &   A Dual control for exploration and exploitation (DCEE)  framework for autonomous search to acquire environment knowledge by exploring the search domain and navigating the search agent closer to the estimated source position.\\ \hline

\citet{Moon2022} & TIGRIS (sampling-based planner) 
   
   Rapidly exploring Information Gathering RIG-tree & A sampling-based planning approach to estimate information gain in larger and higher-dimensional search spaces by prioritizing the exploration of paths with high reward. \\ \hline 

\citet{Ruckin2022} & MCTS with a Deep Reinforcement Learning (RL)  & A RL algorithm for planning in a continuous high-dimensional state and large action spaces. \\ \hline 

\citet{Schmid2022} & SC-aware IPP

Deep 3D SC

PALNet & An Informative Path Planning system integrating a multi-layer mapping Scene Completion (SC) to obtain an SC-aware information gain for fast exploration planning. \\ \hline 

\citet{Tan2022} & Multi-Step DCEE
   
   Stochastic model predictive control (SMPC) 
   
   IPP algorithms & DCEE algorithm for the autonomous search problem of predicting the expected concentration of unknown source locations in the Atmospheric Transport and Dispersion (ATD) model.\\ \hline 

\end{longtable}}
\end{center}


\section{Exploration Path Planning}
\label{sec:robotic_exploration}

Autonomous Exploration Path Planning (EPP) is a high-level decision-making process~\cite{Li2022}. One of the main questions of exploration involves determining where the UAV should go to obtain the greatest possible gain of new information, given the knowledge about the world.

Initially, the exploration problem begins with the principle that the UAV only has information about its own location in the world. Then, it needs to decide where to go next in order to obtain maximum information gain and build a map accurately describing its environment~\cite{Yamauchi1997}. Thus, the decision-making process includes the continuous process of collecting measurements in order to make decisions.

We can observe several exploration strategies, all maximizing information gained (Table \ref{tab:works_EPP}), and different information gain formulations have been proposed to obtain a path as discussed in a recent study by \citet{duberg2022ufoexplorer}. 

\begin{center}
{\footnotesize
\begin{longtable}[H]{p{3cm}p{3.7cm}p{5.5cm}}

\caption{Techniques solving the EPP.} \label{tab:works_EPP} \\

\hline \textbf{Author}& {\textbf{Methods}} & {\textbf{Objective}} \\ \hline 
\endfirsthead
\multicolumn{3}{l}{{\textbf{Continuation - Techniques solving EPP}}} \\
\hline \textbf{Author}& {\textbf{Methods}} & {\textbf{Objective}}  \\ \hline 
\endhead

\hline \multicolumn{3}{l}{{Continued on next page}} \\ \hline
\endfoot

\hline \hline
\endlastfoot

\citet{Bircher2016} \cite{bircher2018receding}  & Receding Horizon 
      
      Next-Best-View Planner (NBVP)
      
      RRT  &  A receding-horizon path planning strategy computes an optimal path in an RRT and only executes the first segment of the branch to the optimal path selected as the highest gain branch. Thus, the planning step is repeated and reinitializes the tree in the next planning iteration on a receding horizon. 
      \\ \hline
 \citet{cieslewski2017rapid}  & Frontier-based planning
 
 NBV method & A fast exploration algorithm to generate velocity commands to quickly reach frontiers in the current field of view. \\ \hline

\citet{Selin2019} & Receding Horizon NBVP with RRT

Frontier exploration

Gaussian Process & 
A receding Horizon Next-Best-View planning combined with Frontier-based exploration is used for a local and global exploration strategy. The potential information gain is estimated using a 360-degree raycasting operation and cubature integration to select the best yaw. From earlier iterations, cached points of the potential information gain are used to interpolate new queries for a Gaussian Process in a continuous domain. Thus, suitably estimating the gain together with its uncertainty.
\\ \hline
      
\citet{meng2017two} & Frontier-based Coverage Planning
   
Fixed-start Open Traveling Salesman Problem (FSOTSP)

SLAM
&  A combination of two-stage planning approaches provides an optimized view planning strategy for volumetric environment model construction without a-prior information concerning the problem of full coverage and the global optimality of the exploration path.\\ \hline

\citet{dang2020graph} & GBPlanner

Rapidly exploring random graph algorithm

Dijkstra's algorithm,

Dynamic Time Warping (DTW) method

Remaining Endurance Time (RET)

Estimated Time of Arrival (ETA)

Model Predictive Control (MPC)  & A graph-based exploration path planner (GBPlanner) ensures efficient and robust exploration of underground environments. It employs a bifurcated local and global planning approach, wherein the local planner efficiently optimizes volumetric gain, combined with a global planner that ensures repositioning of the robot towards frontiers and enables safe auto-homing.
     \\ \hline

\citet{Schmid2020} & RRT*, Truncated Signed Distance Function (TSDF)  & An RRT* inspired online Informative Path Planning algorithm, with a global set of evaluated viewpoints for NBV planning to retain and reuse relevant information efficiently. The information gain considers unmapped volume, and the RRT* expands a single tree of candidate trajectories and rewires after each planning iteration to reduce unnecessary information gain recalculation, allowing the tree to grow continuously with new nodes in the planning iteration.\\ \hline

\citet{batinovic2021multi} & NBVP 

Google Cartographer simultaneous localization

SLAM & A planner for 3D exploration based on detecting a frontier with a clustering algorithm to select the best global point on data obtained by a mapping algorithm. This approach creates an occupancy grid map using Cartographer SLAM and generates an OctoMap.\\ \hline

\citet{lindqvist2021exploration} & Nonlinear Model Predictive Control (NMPC)

E-RRT 
      
& Exploration-RRT (ERRT) framework searches the path considering the potential information gain, the distance traveled, and the movement costs (a receding horizon NMPC).\\ \hline

\citet{kompis2021informed} & Next-Best-View

APF ranking

A*

B-Spline & An informed sampling-based exploration path planner for 3D reconstruction uses the current version of the reconstructed model to propose viewpoints orthogonal to the surface frontiers visible during flight. The APF evaluates an information gain formulation considering exploration and surface quality. The path planner computes a path to the NBV, while considering the current state of the reconstruction and optimizing the path and camera gimbal orientation.\\ \hline

\citet{zhou2021fuel} &  Frontier information structure (FIS)

Greedy exploration strategies

Asymmetric TSP (ATSP) 

Minimum-time B-spline Trajectory &  A fast and autonomous exploration algorithm based on a frontier structure that is incrementally updated for high-frequency exploration planning. Additionally, a hierarchical planning method generates efficient global coverage paths, ensuring safe and agile local maneuvers for high-speed exploration. \\ \hline

\citet{chen2021dual} & Goal-Oriented Control System (GOCS) 
   
   Stochastic Model Predictive Control (SMPC)

   IPP
   
   Isotropic plume model
   & A DCEE control framework combines the exploiting advantages of SMPC and IPP. The exploitation (SMPC) moves the robot towards the predicted source location and exploration (IPP) of an unknown environment to reduce source parameter uncertainties. The cost function is defined as the expected error between the robot’s predicted future position and the predicted future estimated of the new point regarding the information gain of the predicted future measurements. \\ \hline

\citet{respall2021fast} & NBV
      
RRT & A combination of NBV local planning and frontier-based global planning guarantees coverage in unexplored areas. An RRT is used for sampling points and generating collision-free paths. The node with the highest potential gain is selected as the NBV.
Thus, the information gain approach over the estimated time reaches the sample point and the yaw angle.  
  \\ \hline
  
 \citet{duberg2022ufoexplorer} & A multiple-goal A* path generation

R*-tree

UFOMap - volumetric mapping

Pure pursuit path tracking controller & A dense, graph-based planning structure is updated and extended, amortizing the cost of planning. Path planning uses a simple heuristic to guide the robot's movement toward the closest optimal exploration goal and react quickly in larger environments.
   \\ \hline

\citet{Li2022}& DCEE
   
   Bayesian Optimization
   
   MPC
   
   Isotropic plume model
   
   Particle filter  &   A DCEE framework for autonomous search collects information by exploring environment knowledge and plans the robot's next movement to the estimated position using the updated knowledge.\\ \hline 

\citet{ludhiyani2022computationally} & Tailored RRT algorithm and the global set of frontiers divided into three subsets associated with the frontiers planning strategy (Frontiers in Local Neighborhood, 

Remaining Frontiers in 3d Octomap, and Frontier in Current FoV) & A strategy for building accurate maps of large and complex scenarios during high-speed exploration to reduce the dependency on a global set of frontiers. As a result, it increases exploration speed and decreases the computational load of time complexity.\\ \hline

\citet{petit2022tape} & Tether-Aware Path Planning for Exploration (TAPE)
      
TSP
      
RRT-Rope algorithm
      
& A two-level hierarchical architecture. The global frontier-based planning solves a TSP to minimize the distance. The local planning adopts an adjustable decision function whose parameters play on the trade-off between the tether length and the path cost minimization. \\ \hline

\citet{pittol2022loop} &  Coverage area 
      
      Gmapping
      Loop-Aware Exploration Graph (LAEG) & LAEG uses the covered area concept to create a graph in which nodes and edges represent the essential information for exploration. Afterward, the robot incorporates it to support all exploration decisions, not just to relocate globally and avoid predicted loop closures.  \\ \hline

\citet{Schmid2022} & SC-aware IPP
   
   Deep 3D Scene Completion (SC)
   
PALNet & An Informative Path Planning system integrates a multi-layer mapping SC to obtain an SC-aware information gain for fast exploration planning. \\ \hline

\citet{Sun2022} & Frontier detector method based on adaptive RRT & 

Environment structure predictions using previously-acquired information to improve frontier detection efficiency by limiting RRT's sampling space, and generating a non-uniform sampling strategy to solve the over-sampling problem between two adjacent sliding windows. \\ \hline

\citet{Tan2022} &  DCEE

SMPC

IPP

Bayesian Estimation

Bayesian filter & A Multi-Step DCEE (MS-DCEE) framework searches for a path from an estimated location within an unknown environment. It approaches recursive feasibility and convergence analysis of the continuous changes with the estimated location update after each new data, i.e., uncertainty quantification and planning. 
\\ \hline

\citet{zhao2022faep} &  ATSP
   Two-Stage Heading Planning Method

Guided Kinodynamic Path Searching & A fast, autonomous exploration framework improves the frontier's exploration sequence with a low proportion of maneuvers in explored areas. \\ \hline

\end{longtable}}
\end{center}

Robotic exploration has been focused on two main approaches to map unknown environments: exploring space quickly \cite{zhou2021fuel,cieslewski2017rapid,foehn2022alphapilot,petracek2021caves,duberg2022ufoexplorer,dharmadhikari2020motion} and emphasizing accurate reconstruction \cite{song2017online,palazzolo2018effective,kratky2021,Selin2019,xu2021exploration,ludhiyani2022computationally,meng2017two,Schmid2020},  i.e., the trade-offs between cost and information gain. It is important to emphasize that exploration depends on path planning, and the accuracy of the mapping methods depends directly on the localization methods \cite{bourgault2002information}. This paper aims to show how the path planning approach has been adopted in the decision-making processes for exploration.

Table \ref{tab:works_EPP} presents some existing proposals for EPP. Analyzing these works, we can observe that, in order to solve EPP, different approaches have been developed, including search-based methods \cite{Yamauchi1997,heng2015efficient,deng2020robotic,zhong2021information,julia2012comparison,dai2020fast,batinovic2021multi,wang2019efficient,schneider2016fast} and sampling-based methods \cite{cao2021tare,Zhu2021dsvp,maciel2019online,xu2021exploration,respall2021fast,wang2019efficient,oleynikova2018safe,foehn2022alphapilot,Papachristos2017uncertainty,Bircher2016}. 

Random sampling-based methods are based on an RRT~\cite{Lavalle2000} or similar techniques~\cite{bhattacharya2010search}. In general, the methods include the objectives of sampling the next–best–view~\cite{Bircher2016} and graph-based exploration~\cite{dang2020graph}, which have proven effective in tunnel-like environments, although they neglect some areas. Moreover, when a rewarding gain is assigned, the methods may also show greedy behavior similar to that of information theory methods \cite{petit2022tape}.

Some approaches combine the search-based and sampling-based methods \cite{zhou2021fuel,song2017online,Charrow2015,Selin2019,meng2017two,yang2021graph,respall2021fast,cao2021tare,Zhu2021dsvp}, while another focus covers multi-objective planning \cite{papachristos2019autonomous,Papachristos2017uncertainty}. Recently, in Exploration Path Planning, some Deep Learning methods have been used  \cite{reinhart2020learning,Li2022,maciel2019online,qi2023multi,niroui2019deep}. Path planning exploration involving multiple robots has also been proposed \cite{batinovic2021multi,zhou2022racer,li2022multi,Song2022multi,ivic2022multi,fu2022multi,qi2023multi}.

Frontier-based methods address the boundaries separating known space from unknown space~\cite{Yamauchi1997}. The frontier-based planning finds the frontiers with the highest information gain according to the map established and the robot's movement \cite{dornhege2013frontier,gao2018improved}.

The frontiers-based methods are able to quickly explore the entire environment, guaranteeing global coverage. However, the process of finding and describing boundaries can be computationally expensive in large volumetric maps~\cite{zhao2022faep}. Instead, a utility function can be employed to iteratively determine the best frontier until there are no more unknown spaces available~\cite{naazare2022online}.

In the classical frontier-based approach, \cite{Yamauchi1997}, the boundaries are identified as known-free voxels in close proximity to unknown ones. They are then grouped into clusters to select known points to be visited. However, the extracted information is too coarse to perform fine-grained decision-making~\cite{zhou2021fuel}. 

The frontier-based approach can use the RRT method to uniformly expand the search into the surrounding area \cite{Umari2017}\cite{Sun2022}\cite{Bircher2016}. Compared to the frontiers-based approach, the NBV approach presents local solutions for areas that have been sampled randomly. NBV aims to determine the best points for obtaining the largest quantity of information \cite{maboudi2022review}. 

 As an extension to \cite{Yamauchi1997}, \citet{cieslewski2017rapid} proposes an algorithm for exploration at high speeds. In the NBV method, the target frontier within the field of view (FoV) is selected for each decision using a reactive behavior of the algorithm in order to incorporate new information faster. The planning method considers the next point requiring the minimum modification of the UAV's movement, such that its speed is not altered. This method minimizes speed changes and maintains a consistently high flight speed. The main limitation of the NBV method is the map extension process, as it does not accurately explore a previously explored area \cite{Zhu2021dsvp}.

\citet{Zhu2021dsvp} proposes a dynamically extended frontier detector to give the highest priority to path planning. 
While searching, an extended RRT is used to guide the expansion of RRT with hybrid frontiers. It should be mentioned that the frontiers are extracted from sensors associated with RRT nodes.
The utility function computes each branch's gain in the tree by adopting Dynamic Time Warping (DTW) in order to identify the similarity between the actual and selected branches, which reflects the direction being explored. However, this method presents overlapping sampling areas that can cause the UAV not to observe newly explored areas beyond the selected boundaries or maintain them in areas with complex obstacles. \citet{Sun2022} proposes an adaptive Rapidly-exploring Random algorithm based on a sliding window of varying size to enhance the sampling rate to a satisfactory and successful extent. Further, to solve the over-sampling problem of \cite{Zhu2021dsvp}, a non-uniform sampling strategy is proposed. 
The utility function is the same as the one used in \cite{Zhu2021dsvp}.

The sampling-based strategies eliminate the computation of the global set of frontiers. Instead, these strategies sample poses from the known free space and evaluate them based on a utility function in order to identify the next potential pose for exploration. As this approach does not depend on the global frontiers, its computational cost should be less expensive compared to frontier-based approaches \cite{ludhiyani2022computationally}.

Sampling-based approaches allow for a wide range of information gain formulation and generate viewpoints randomly to explore the space \cite{naazare2022online}. State-of-the-art sampling-based exploration methods adopt the receding horizon approach~\cite{Bircher2016,Schmid2020,Selin2019}.  

The first NBV planner was presented by \cite{Bircher2016}. Herein, the authors proposed a receding-horizon path planning strategy that computes an optimal path in a rapidly exploring random tree (RRT). It only executes the first step of the optimal path selected as the best maximum information gain branch; the rest is used to reinitialize the tree in the next planning iteration on a receding horizon. The information gain is defined as the sum of the unmapped volume able to be explored (i.e., the space observed by sensors and not occupied by obstacles). The planning frontier used was the robot's field of view  (FoV), whose exploration gain exceeded the preset threshold. An extension of the \cite{Bircher2016} was proposed by \cite{bircher2018receding} for surface inspection. 

In \citet{Bircher2016}, an efficient global route is ignored in the greedy strategy, resulting in unnecessary re-visitation and reducing the overall efficiency \cite{yan2022mui}. Considering this behavior, a two-stage heuristic information gain-based next-view planning algorithm was proposed by \cite{meng2017two}. The approach is based on coverage planning with a Fixed Start Open Traveling Salesman Problem (FSOTSP). The frontier viewpoints are first sampled in an optimal open exploration sequence without a priori knowledge, followed by finding the global shortest tour through all viewpoints. The utility function is based on Euclidean distance and an information gain-based heuristic cost. Furthermore, the cost function is reversely proportional to information gain, which represents a significant penalty applied to the traveling distance.

A Belief Uncertainty-aware Receding Horizon Exploration and Mapping path planning strategy was proposed by \cite{Papachristos2019localization}. The strategy has two steps to solve the EPP problem: exploration and uncertainty optimization. At the exploration level, the path planner explores the environment using an optimized sequence of points to be visited, prioritizing the discovery of new and unknown areas and improving areas with uncertainty. Instead of the UAV blindly moving to the next best point, the uncertainty optimization level evaluates alternative paths that can lead to increased localization and mapping accuracy. Then, the path planner acquires information for each of the levels and repeats the process in a receding horizon while considering the complete progressive exploration of the environment.

\citet{zhou2021fuel} also proposes a fast exploration framework (FUEL - Fast UAV ExpLoration) based on incremental frontier structure (FIS) and hierarchical planning in order to obtain an efficient global route. The authors improve the information extracted from the frontier, producing the best planning in three steps. First, it finds an optimal global exploration to accumulate information about the environment using a TSP. This is done in order to compute the shortest path while visiting candidate viewpoints. Then, it refines a local segment, improving the exploration rate, and generates a feasible and minimum-time trajectory. This method makes decisions quickly and provides a smooth and high-speed exploration trajectory in complex environments. However, the limitation of this algorithm is that it does not take into account the state estimation uncertainty in the exploration efficiency. Revisitation during the exploration process will be achieved by works in~\cite{zhao2022faep,yan2022mui}. Since the method focuses exclusively on path planning for the sampled areas rather than prioritizing unexplored unknown areas.

Therefore, based on FUEL, \citet{zhao2022faep} proposes to improve the reasonable frontiers exploration sequence by considering the influence of global exploration and the stability of path searching in individual environments. The exploration priority should be higher for a small unknown area on the path or for a frontier closer to the boundary of the exploration area, while avoiding unnecessary maneuvers. A guided EPP method is adopted to guide the direction of kinodynamic path searching.
The authors also consider Adaptive Dynamic Planning, which maintains stable, high speed during flight. It selects a future starting point from the current planning, avoiding low-speed or stop-and-go maneuvers.

From a path planning perspective, there is a limitation in the tree structure for searching between two nodes, whereby the search to a node is restricted by its connection to the root node~\cite{duberg2022ufoexplorer}. This means that the search is limited to the nodes connected to the root node, making it impossible to search for a path between two nodes not connected to the root node. This limitation can be overcome by using a graph structure, which allows for searching between any two nodes. Therefore, some works have focused on the graph-based exploration approach \cite{Papachristos2017uncertainty,dang2020graph,duberg2022ufoexplorer}. 

As an extension of \cite{Bircher2016} and \cite{nieuwenhuisen2019search}, a two-stage path planning architecture consisting of a rapid tree-based local exploration layer and a graph-based global planning layer was proposed by \citet{dang2019explore}. This greedy search strategy adopts the RRT* algorithm in the local layer, exploiting the local space surrounding the robot to facilitate a faster response. For the global planner, the local search graph is used to look for the nearest neighbors of each new node in order to find a viable solution without collisions and at a lower cost. In addition, the DTW similarity metric is applied to decrease the search, thereby decreasing the number of potential paths.

A tailored, graph‐based EPP strategy through a local and global planning architecture was proposed by~\cite{dang2020graph}. The local planning utilizes the rapidly exploring random graph algorithm to optimize exploration gain within the local subspace, while simultaneously 
avoiding obstacles, considering constraints and the environment, as well as the dynamic limitations of the robot. 
In situations where the robot must either be replaced in a previously identified frontier of the exploration space (unexplored areas) or return home, the global Rapidly-exploring Random Graph (RRG) works together with local planning to incrementally build a sparse global graph.

An informed, sampling-based exploration path planner was proposed by \citet{kompis2021informed} using a surface frontier instead of the original volumetric frontier. The approach is divided into two parallel processes: an NBV evaluator and a trajectory planner. The NBV evaluator process evaluates the global point of a subset of sample viewpoints received by an artificial potential field (APF) ranking scheme, which discards points with low information gain. The APF predicts the information gain of the viewpoint based on its location and surroundings. 
The trajectory planner uses two-stage planning, including A* path search and B-Spline trajectory optimization. As a result, this informed sampling-based approach most often leads to the highest proposed viewpoints with high information gain, while also considering additional degrees of freedom during sampling, such as the yaw and pitch of the sensor.

An approach based on a dense undirected graph keeping all information about the known areas of the map is proposed in~\cite{duberg2022ufoexplorer}. Then, the new path is calculated before the first step has been fully executed by matching the path and map updates, amortizing the cost of planning.  By differentiating between nodes with and without information gain, the search limits the number of steps in which information gain must be calculated. The gain information and yaw are calculated for every newly added node. 


An approach to safe autonomous exploration and map building based on the Bayesian Optimization method was proposed by \cite{francis2019occupancy}. This method finds ideal continuous paths, rather than discrete detection locations satisfying UAV movement and safety restrictions. By balancing the reward function and risk associated with each path, the optimizer minimizes the number of computationally expensive function evaluations. The authors adopted a Restricted Bayesian Optimization, which is a safe exploration method to learn a Mutual Information (MI) objective. While the acquisition function optimizes continuous paths, it uses only limited parameterization in practice, such as quadratic or cubic splines.

The proposed Learning-Based Path Planning Exploration (LBPlanner) achieves an exploratory behavior with reduced computational cost and without the need for a consistent map \cite{reinhart2020learning}. The graph-based path planner, GBPlanner \cite{dang2019explore}, is used as an expert planner to provide training data. The LBPlanner assumes an always-available volumetric representation of the environment and only applies a sliding window of observations for its training and inference steps. This approach does not consider the relationship between the previous observations and the final results, nor does it provide a map. 

Given the challenges of limited information and occlusions in the exploration planning, \citet{Schmid2022} proposed an SC-Explorer framework that uses the mapping approach with non-blocking raycasting and an SC-aware information gain in the path planning. 
Based on the incoming depth images while fusing the predictions spatially over time and combining them with the fused measured data into a hierarchical multi-layer map, 3D scene completions are predicted. The robot’s SC-aware IPP receives information in detail from the map. 
The gain formulation leads to the robot implicitly adapting to the type of environment and emphasizing the predicted areas, while still retaining a pure exploration aspect for areas beyond those predicted or for those that are unidentified.

In order to offer a better autonomous search strategy, \citet{Li2022} adopted the concept of Dual Control for Exploration and Exploitation (DCEE) \cite{chen2021dual} in an information-oriented IPP strategy combined with an SMPC strategy. Together, they developed a concurrent learning exploration strategy to collect more knowledge of the environment to reduce estimation uncertainty. In the decision-making process and the environment estimation, environment-learning and IPP algorithms were tailored to acquire information by exploring the search domain and planning the path closest to the estimated source position.
\citet{Tan2022} reformulated the autonomous search exploration proposed by \cite{Li2022} as a Multi-Step DCEE (MS-DCEE) framework to obtain the recursive feasibility and convergence properties. Thus, the MS-DCEE probes and learns to reduce the variance estimated by balancing exploitation and exploration.

\section{Problem Formulation}
\label{sec:Problem_formulation}

This section presents a brief clarification about the informative path planning problem introduced in Section (\ref{sec:Problem_formulation_IPP}) and the exploration path planning problem in Section (\ref{sec:Problem_formulation_EPP}).

\subsection{Informative Path Planning Problem}
\label{sec:Problem_formulation_IPP}

This section briefly describes two widely-adopted representations of the Informative Path Planning problem. In \textit{Information gain maximization approach 1} (Sec. \ref{sec:IPPApproach1}), the focus is on maximizing information gain, I($\cdot$). In \textit{Information gain maximization approach 2}, the focus is on the rate of the information gain, which applies to tasks that require comparison in the quality of the information of the paths in different scales of time and length. It is important to note that several strategies can be adopted to solve the objective function, such as Entropy~\cite{Ercolani2022}, Kullback-Leibler Divergence (KLD)~\cite{Ercolani2022}, Lévy flight~\cite{Ercolani2022}, Upper Confidence Bound (UCB)~\cite{Meera2019}, Shannon entropy~\cite{Popovic2020_IPP,Popovic2020informative}, and Rényi entropy~\cite{Popovic2020_IPP}, among others.

\subsubsection{Information gain maximization approach 1}
\label{sec:IPPApproach1}

The general IPP problem is formulated to find the optimal trajectory \(T^{*}\) of a UAV to maximize knowledge in a specific environment $\varepsilon$. In other words, this trajectory should pass through the space of all possible continuous paths \(\mathbb{T}\) to maximize an information gain measure $I$($T$) subject to a budget constraint \(B\), as shown in Equations. 

\begin{equation}
\label{eq:IPP1_1}
        T^{*}=\arg\max_{T \in \mathbb{T}}  I \left(\mathrm{Measure}(T)\right),
\end{equation}

\begin{equation}
\label{eq:IPP1_2}
         C(T) \le (B),
\end{equation}

 \noindent where $T$ is a path defined as a set of waypoints in \(\mathbb{R}\)$^3$. 
 Measure ($\cdot$) provides the finite set of measurements $x_j$ along path $T$. The \(\mathrm{I}\)($\cdot$) defines the information objective quantifying the utility of these measurements.
The cost function for the trajectory is provided by $\mathrm{C}$($\cdot$), which cannot exceed the budget. The budget constraint is defined by the application problem; this could be mission time, path length, etc. This approach is used in \cite{Velasco2020,Moon2022,McCammon2021,Hollinger2014,singh2009efficient,Popovic2020_IPP,Bostrom2019,Batinovic2022shadowcasting}. The Table \ref{tab:approach_1} shows further detail about these works.

\begin{center}
{\footnotesize
\begin{longtable}[H]{p{1.7cm}p{1.7cm}p{1.7cm}p{3cm}p{2cm}p{1.5cm}}

\caption{Information gain maximization approach 1 - IPP} \label{tab:approach_1} \\

\hline \textbf{Reference}& {\textbf{Approach}} & {\textbf{Optimize}}& {\textbf{I($\cdot$) }}& {\textbf{Measure($\cdot$)}}& {\textbf{Test}}\\ \hline 
\endfirsthead
\multicolumn{6}{l}{{\bfseries \tablename\ \thetable{} -- Cont.Information gain maximization approach 1 - IPP}} \\
\hline \textbf{Reference} & {\textbf{Approach}} & {\textbf{Optimize}}& {\textbf{I($\cdot$) }}& {\textbf{Measure($\cdot$)}}& {\textbf{Test}} \\ \hline 
\endhead

\hline \multicolumn{6}{l}{{Continued on next page}} \\ \hline
\endfoot

\hline \hline
\endlastfoot

    \citet{singh2009efficient}   &   Greedy algorithm and Non-myopic adaptive    &   Distance   &  Padded decomposition   &  Set of observations locations   & 3D  Simulated    \\ \hline
    \citet{Hollinger2014}               &  RIG-roadmap     &  Mission time and Fuel    &   
    Modular (cost), time-varying modular (time),  and submodular information objective (intersection of trajectories)    &  Time    &  2D Simulated      \\ \hline
    \citet{Bostrom2019}      & Finite-horizon     &  Tracking performance    & Stochastic target trajectory and measurement noise    &  Covariance matrix of the target state estimate    & 2D Simulated\\ \hline
    \citet{Popovic2020_IPP}   & Fixed-horizon     & Rényi entropy      & Smooth minimum-snap dynamics     &  A constant sensor frequency and robot speed   &  3D Simulated \\ \hline
   \citet{Velasco2020}       &  Finite-horizon     &  Computational cost    &   Adaptive Cascaded Local Optimal Planning (ACLOP) &  Variance of Gas Distribution Models (GDM) using Kernel DM+V   &  2D Simulated\\ \hline
   \citet{McCammon2021}      &  Greedy Coverage      &  Distance    & Optimal Schedule     &  Length   &  2D Simulated     \\ \hline
    \citet{Moon2022}       &  RIG - Tree, Tree-based Information Gathering using Informed Sampling     &  Exploration of paths with high reward    &  Prune(·) function and an upper bound on reward   &  The state, path cost, and path information   &  3D Simulated   \\ \hline
    \citet{Batinovic2022shadowcasting}    &   Receding Horizon     &  Best path    &  
     The volume of the unknown space covered is weighted with a negative exponential of the cost to travel along the path      
     &   Receding Horizon Next-Best-View (RH-NBVP)   &   3D Simulated  

\end{longtable}}
\end{center}

\subsubsection{Information gain maximization approach 2}
\label{sec:IPP_Approach2}

In contrast to \textit{Information gain maximization approach 1}, it is formulated in the space of trajectories $T$. According to \citet{Popovic2017multi}, the paths can be described as time functions and obtain measurements with a constant frequency, while also considering the practical application needs. The aim is to find an optimal continuous trajectory $T^*$ in the space of all feasible trajectories \(\mathbb{T}\) for the maximum information gain collected about the environment, while still accounting for budget constraints.

\begin{equation}
\label{eq:IPP2_1}
        T^{*}=\arg\max_{T \in \mathbb{T}} \frac{ \mathrm{I} \left(Measure(T)\right)}{C(T)},
\end{equation}

\begin{equation}
\label{eq:IPP2_2}
         C(T) \le (B),
\end{equation}

\noindent where $C$: \(\mathbb{T}\) $\rightarrow $  \(\mathbb{R}^{+}\) plans an action sequence to its associated
corresponding execution cost, which cannot exceed a predefined budget, as indicated by $B$. $I$ ($\cdot$) quantifies the utility of the acquired measurements by obtaining the informative objective.
The function $Measure$($\cdot$) obtains a finite set of measurements along trajectory $T$ in an environment \(\mathbb{R}\)$^3$.

The function $Measure$ ($\cdot$) obtains measurements along the path $T$. A utility rate maximization in the Eq. \ref{eq:IPP2_1} enables comparing the values of paths over different time scales, instead of only maximizing utility $I$ ($\cdot$). \textit{Approach 2} shows a generic utility 
that expresses the expected reduction in the map’s uncertainty \cite{Popovic2017online}.

The costs $C$($T$) of an action sequence $T$  = ($T_1$, ..., $T_n$) of length $n$ can be defined by the total travel time~\cite{Ruckin2022}:

\begin{equation}
\label{eq:IPP2_3}
C(T) = \sum_{i-1}^{n-1} c(T_i, T_{i+1}).
\end{equation}

\citet{Zhu2021} considered the restriction of the Eq. \ref{eq:IPP2_4}, where Collision($\cdot$) returns the collision points of the trajectory. Hence, by enforcing constraints Eq. \ref{eq:IPP2_3} and \ref{eq:IPP2_4}, the planned trajectory should respect the time budget and be collision-free in the environment. 

\begin{equation}
\label{eq:IPP2_4}
\mathrm{Collision}(T) = 0.
\end{equation}

This approach can also be found in \cite{Popovic2017multi,Popovic2017online,Popovic2020informative,Zhu2021,Meera2019}. Table \ref{tab:approach_2} shows further detail about these works.

\begin{center}
{\footnotesize
\begin{longtable}[H]{p{1.7cm}p{1.5cm}p{1.6cm}p{2cm}p{2.2cm}p{1.2cm}p{1.5cm}}

\caption{Information gain maximization approach 2 - IPP} \label{tab:approach_2} \\

\hline{\textbf{Reference}} & {\textbf{Approach}} & {\textbf{Optimize}}& {\textbf{ I($\cdot$)}}& {\textbf{Measure($\cdot$)}}& {\textbf{C($\cdot$)}}& {\textbf{Test}}\endfirsthead\\ \hline

\hline {\textbf{Reference}} & {\textbf{Approach}} & {\textbf{Optimize}} & {\textbf{ I($\cdot$)}} & {\textbf{Measure($\cdot$)}}& {\textbf{C($\cdot$)}}& {\textbf{Test}}\endhead \\ \hline

\hline \multicolumn{7}{l}{{Continued on next page}} \\ \hline
\endfoot

\hline \hline
\endlastfoot

\citet{Popovic2017multi}   & Fixed-horizon    &  Trajectory & 3-D grid search and evolutionary optimization  &  An altitude-dependent Gaussian sensor noise model   & Travel time    &  3D Simulated and Real  \\ \hline
\citet{Popovic2017online}    & 
Fixed-horizon  & Polynomial path  &  Evolutionary strategy    &  Weed classifier output   &  Travel time    &  3D Simulated and Real   \\ \hline
\citet{Popovic2020informative}    &  Fixed-horizon    &  The coarse grid search   &   Shannon’s entropy  &  Discrete or continuous target variable of interest  &  Travel time   & 3D Simulated and Real \\ \hline
\citet{Zhu2021}& Receding horizon manner    & Continuous trajectory   &  Sequential greedy search and 3D Euclidean Signed Distance Field
(ESDF)    &  A constant sensor measurement frequency    &  Travel time    &   3D Simulated   \\ \hline
\citet{Meera2019}&  Adaptive, non-myopic, and continuous  &  Uncertainty reduction   & Evolutionary optimizer CMA-ES  &   Probabilistic sensor models for data fusion, with constant-time measurement updates    &  Travel time and Collision cost & 3D Simulated   \\ 

\end{longtable}}
\end{center}

\subsection{Exploration Path Planning Problem}
\label{sec:Problem_formulation_EPP}

For \textit{Exploration Path Planning} of a previously unknown environment, the objective is to autonomously build the spatiotemporal distribution of a quantity of interest $\mathcal{M}: \mathbb{R}^3 \times \mathbb{R}^+ \rightarrow \mathbb{R}$ from the sensor's measurements and localization poses. The vector of coordinates identifying a robot's position at time $t$ can be defined by $z_t =[x_t,y_t,z_t,\psi_t]^T$.

\subsubsection{World representation} The map $\mathcal{M}$ consists of unknown (${m}_{un}$), occupied (${m}_{occ}$), and free (${m}_{free}$) spaces. The area of interest is represented by a 3D grid, $\mathcal{V} \subset \mathbb{R}^3$. Each cell ${v}$ has a robot state in the map  ${M}$ at time $t$, $M_t(v): \mathbb{R}^3 \rightarrow \mathcal{M}$, and the observed space at time, $\hat{\mathcal{V}}_t$ \{${v}$ $\epsilon$ $\mathcal{V}$ $|$ $M_t(v) \neq {m}_{un}$\} \cite{Schmid2022}.

\subsubsection{Volumetric exploration planning approach} 

The volumetric exploration planning approach maximizes the explored area given the current knowledge of known and unknown space. It requires finding a feasible path to navigate the robot by identifying a set of viewpoints in a 3D area and distinguishing them into either \{$\mathcal{V}_{res},\mathcal{V}_{occ},\mathcal{V}_{free}$\} $ \subseteq M$.
Thus, exploration is considered complete if the explored volume $ \mathcal{V}_{occ} \cup  \mathcal{V}_{free} = \mathcal{V} \setminus \mathcal{V}_{res}$, where $V$ is the volume within the bounds of the unknown space to be explored \cite{Papachristos2019localization,dang2020graph}. 

Some works propose a solution for EPP using local and global exploration \cite{dang2019explore,duberg2022ufoexplorer,kompis2021informed,dang2020graph}. Local exploration focuses on exploiting the local space around the robot, while global exploration executes the planning in global space.

\textbf{\textit{Local Exploration}} finds an admissible path that is collision-free and safety-aware, maximizing the exploration gain in unmapped areas \cite{dang2019explore}, 

\begin{equation}
\label{eq:localexploration}
\begin{split}    
E_g(\sigma_i)= 
    e^{-\gamma_\mathcal{S} {\mathcal{S}(\gamma_i,\gamma_{ref})}}  \\ 
    \sum_{j=1}^{m_i}V_g(\mathcal{V}_{j}^{i})e^{-\gamma_\mathcal{D} {\mathcal{D}(\mathcal{V}_{1}^{i},\mathcal{V}_{j}^{i})} },
\end{split}
\end{equation}

\noindent where $E_g$ and $V_g$ represent Exploration Gain and Volumetric Gain, respectively, $\gamma_i$ is a path from the root search to all leaf nodes, $i=1,...,n$, and $\mathcal{V}_{j}^{i} = 1,...,m_i$ are nodes along this path. $V_g$ is the total unknown volume expected that can be perceived from the measurements sensor in the robot position node $\mathcal{V}_{j}^{i}$. A penalization factor $\gamma_\mathcal{D}$ is adopted to penalize the informative paths with longer distances. The euclidean distance from each node is given by $\mathcal{D}(\mathcal{V}_{1}^{i},\mathcal{V}_{j}^{i})$. A similarity $\mathcal{S}(\gamma_i,\gamma_{ref})$ between a path $\gamma_i,$ and a pseudo-straight path $\gamma_{ref}$ can be aligned with the currently estimated exploration direction~\cite{dang2019explore,dang2020graph}. 

\textbf{\textit{Global Exploration}} provides a path to unexplored areas of the map when the local exploration has explored the entirety of its area and a path for returning the robot home safely, where $G$ represents the Global Exploration Gain as:

\begin{equation}
\label{eq:globalexploration}
\begin{split}    
    G_{G}(\mathcal{V}_{G,i}^{\mathcal{F}})= \\
    \mathcal{T} (\mathcal{V}_{G,cur},\mathcal{V}_{G,i}^{\mathcal{F}})
    V_g(\mathcal{V}_{G,i}^{\mathcal{F}})e^{-\epsilon_\mathcal{D} {\mathcal{V}_{\mathcal{D},cur},(\mathcal{V}_{G,i}^{\mathcal{F}})} },
\end{split}
\end{equation}

\noindent where $\mathcal{V}_{G,cur}$ is the current robot pose in the global graph,  $\mathcal{F}=\mathcal{V}_{G,i}^{\mathcal{F}}$ denotes the set of available paths with high volumetric gain, and the estimated remaining time $\mathcal{T} (\mathcal{V}_{G,cur},\mathcal{V}_{G,i}^{\mathcal{F}})$ is used to explore the available paths for the planner in order to make a selection~\cite{dang2019explore,dang2020graph}.

\section{Methods for Decision-Making-based Path Planning}
\label{sec:Approaches}

In addition to the problem presented previously, this section presents state-of-the-art methods that have been used for solving both exploration and informative path planning. The following methods include the Markov Decision Process \cite{puterman2014markov,ferguson2004focussed,Achour2010,manjanna2022scalable}, Partially Observable Markov Decision Process \cite{Marchant2014,Marchant2012,Fernandez2022,qiming2019modeling,Nguyen2019,Hollinger2009,Morere2017,Salhotra2021,Silver2010,Choudhury2020,carden2022improved,farhi2022bayesian}, Gaussian Process Regression \cite{vivaldini2016,vivaldini2019uav,xiao2022nonmyopic,Brouwer2021,Popovic2017multi,marchant2014bayesian,Morere2017,Hitz2017,Popovic2020informative,Zhu2021,Gao2022}, and Bayesian Optimization \cite{vivaldini2016,vivaldini2019uav,Blanchard2022,Gao2022,Li2022,Ghassemi2022,Yifei2022,Morere2017,francis2019occupancy,Meera2019,marchant2014bayesian,Popovic2020_IPP,Tan2022}.

\subsection{Markov Decision Process for Path Planning}
\label{sec:MDP_pathplanning}

The Markov Decision Process (MDP) is a mathematical formalism for modeling a wide of decision-making problems, in which the environment is fully observable. It can be used as a stochastic decision-making model for finding an optimal strategy. In the decision-making process, the robot makes decisions in the time $t=1,2,...,n$. For each state, the current observation and next action are used. Reward and transition to the next state depend on the action taken and the state in which it was taken \cite{puterman2014markov,ferguson2004focussed,Achour2010}. 

\subsubsection{Definition}
\label{sec:MDP_pathplanning_def} 

An MDP is specified by a tuple ($\mathcal{ S, A, T , R}$), where $S$ is a finite set of state space ${s_1, s_2, ..., s_n}$ and $A$ is a set of possible actions ${a_1, a_2, ..., a_n}$ indicated by two properties.  

The first property includes the probability of transitioning a state $s'$ depending only on the previous state $s$ and action $a$, where $\mathcal{T}: S  \times A \times S \rightarrow  [0,1] $ is a probabilistic transition function. 
That is to say, $\mathcal{P_r} (s'|, s, a)$ is the probability that action $a$ in the state $s$ at time $t$ will lead to state $s'$ at time $t+1$, where the action $a$ $\epsilon$ $A $ and  $s$ $\epsilon$ $S $.

The second property is the reward function $R: S  \times A  \rightarrow  \mathbb{R} $, which assigns a reward for each transition $R (s', s, a)$ from state $s$, the action $a$, and state $s'$. The reward at time $t$ is a function of the current state and action.

\subsubsection{Policy}

A policy describes the behavior of an agent as a set of instructions that the agent can use to determine what action to take in a given state. 
A \textit{deterministic policy} $\pi: S \rightarrow A$ includes the mapping from any given state space to an action, whereas a \textit{randomized policy} $\pi: S \rightarrow \Delta ^{A}$ is mapping from any given state to a distribution over actions, where $\pi_t$ at time $t$ indicates that the robot must execute the action $a_t=\pi_t(s)$ if the current state $s_t = s$.  In a \textit{stationary policy}, $\pi$ represents a mapping action to be taken independent of the time, i.e.,  $\pi_t$ = $\pi$ for all rounds $t$ = 1,2,... etc.

A \textit{non-stationary policy} $\pi(s_t)$ represents the action at time $t$ as a function of the current state $s_t$ \cite{kaelbling1998planning}.

\subsubsection{Optimization objective function}
 The Markov Decision Process evaluates the trade-off between an immediate reward and future rewards of sequential decisions, as well as the need for planning ahead. 
Calculating the optimal value function is a standard way to find an optimal policy, with the objective of maximizing the cumulative function of the rewards.

\textit{Finite Horizon}: finds a sequential decision policy $\pi$ that maximizes the expected reward:

\begin{equation}
\label{eq:MDP_FH}
    \mathbb{R}\left [ \sum_{t= 1}^{T} \gamma^{-1} r_t| s_1   \right ],    0\leq \gamma \leq 1.
\end{equation}

The actions are in a finite horizon for $t= 1, ..., T $. The total reward criterion maximizes the expected total rewards in an episode of length $T$.

\textit{Infinite Horizon}: Expected total discounted reward criteria maximizes the expected discounted sum of rewards. 

\begin{equation}
\label{eq:MDP_IH_etd}
   \lim_{T\rightarrow\infty } \mathbb{R}\left [ \sum_{t= 1}^{T} \gamma^{-1} r_t| s_1   \right ],    0 \leq \gamma \leq 1,
\end{equation}

\noindent where $r_t$ is the immediate reward in time $t$ and $ 0 \leq \gamma < 1$  is a discount factor.
The expected total reward criteria maximizes the ergodic expected reward, as: 

\begin{equation}
\label{eq:MDP_IH_etr}
    \lim_{T\rightarrow\infty }\mathbb{R}\left [ \sum_{t= 1}^{T} r_t| s_1   \right ],    0\leq \gamma \leq 1.
\end{equation}

\subsection{Partially Observable Markov Decision Process - POMDP}

There are numerous works that apply the Partially Observable Markov Decision Process (POMDP) to exploration or Informative Path Planning, such as  \cite{Marchant2014,Fernandez2022,hoerger2021line,jin2020sample,qiming2019modeling,Nguyen2019,nguyen2019trackerbots,Choudhury2020,Hollinger2009,Morere2017,Salhotra2021,Silver2010,golowich2022planning,xiao2019online,reinhart2020learning}.
A survey of the POMDP approaches and a discussion of their strengths and drawbacks are presented by \cite{ross2008online,kurniawati2022partially}. \cite{golowich2022planning} presents an understanding of the gap between the theory and practice of POMDPs. We can compare the \textit{Informative and Exploration Path Planning problem} to the problem of the optimal calculation of actions under uncertainty. The POMDP is an efficient and optimal approach based on previous actions and observations to determine the states and maximize its reward \cite{nguyen2019trackerbots}. 

\subsubsection{Definition}
The POMDP is a general framework for a sequential decision problem in a partially observable environment \cite{ross2008online,Morere2017}. POMDP is similar to the standard MDP, except that instead of the state being observable, there are partial and/or noisy observations about the current state, even when it cannot precisely observe the state of its environment \cite{kaelbling1998planning}.

A POMDP consists of a 6-tuple ($\mathcal{ S, A, T, R, Z, O }$), where $\mathcal{S}$, $\mathcal{A}$, $\mathcal{T}$, and 
$\mathcal{R}$ have the same meaning as in an MDP (Section \ref{sec:MDP_pathplanning_def}). $\mathcal{Z}$  denotes the observation space, i.e., the set of all possible observations. $\mathcal{O} : \mathcal{S} \times \mathcal{A} \times \mathcal{Z} \rightarrow  [0, 1] $ denotes the observation function, where $\mathcal{O}(s', a, z) = \mathcal{P_r}(z|a, s')$ gives the probability of observing $z$ in state $s' \epsilon \mathcal{S}$ in relation to the action  $a \epsilon \mathcal{A}$.
\subsubsection{Belief state}
In an environment where the state is not directly observable due to partial observability, the agent has incomplete information about the current state through observation $z \epsilon \mathcal{Z}$. Thus, it should consider the complete history of actions and observations up to the current time $t$ in order to decide which action to perform:  

\begin{equation}
\label{eq:pomdp_h}
    h_t = {a_0, z_1, ..., z_{t-1}, a_{t-1},z_t}.
\end{equation}

This explicit representation is typically memory expensive, so rather than employing it, one may summarize all relevant previous information in a probability distribution over the state space $\mathcal{S}$, which is denoted as the belief state $b$. 

\begin{equation}
    \label{eq:pomdp_b}
     b_t(s) = P_r (s_t = s |h_t = h), \forall s \epsilon \mathcal{S},
\end{equation}

\noindent where $b_t$ $\epsilon$ $\mathcal{B}$ and $\mathcal{B}$ is the set of all possible belief states \cite{ross2008online}.

The agent infers the best action $a$ $\epsilon$ $\mathcal{A}$ to execute from the current belief $b$ and can improve its knowledge without additional apriori information of the environment. Thus, the belief state $b_t$ at $t+1$ can be updated according to:

\begin{equation}
\label{eq:pomdp_bt}
b_{o}^{a}(s')=\frac{Pr(o|s',a) \sum_{s \epsilon \mathcal{S}}(s'|s,a)b(s)}{ \sum_{s' \epsilon \mathcal{S} } Pr(o|s',a)\sum_{s \epsilon \mathcal{S}}Pr(s'|s,a)b(s)}.
\end{equation}

\subsubsection{Policy}
As in an MDP, solving a POMDP problem means finding an optimal policy $\pi^*$ that optimizes the value function $V^{\pi}$. The POMDP's policy is defined over the belief state $d$ as a mapping $\pi^{*} \mathcal{B}: \rightarrow \mathcal{A}$ from beliefs to actions \cite{kurniawati2022partially,ross2008online}. 
In robotics, one of the most used objective functions is the discounted reward-based POMDP. The value function $V^{\pi}$ can be defined as the cumulative expected discounted rewards by a policy $\pi$ from a certain belief state $b$: 

\begin{equation}
\label{eq:pomdp_p1}
V^{\pi}(b)= \mathbb{E}_{\pi}\left [ \sum_{t=0 }^{\infty} \gamma^\mathcal{R} (b_t,\pi(b_t)) \mid b_0 =b  \right ].
\end{equation}

\textit{\textbf{Discrete state space}}

Equation \ref{eq:pomdp_p2} expresses the expected reward of executing an action in belief state b for the discrete state space.

\begin{equation}
\label{eq:pomdp_p2}
    \mathcal{R} (b_t,\pi(b_t) = a) = \sum_{s \epsilon \mathcal{S}}^{} \mathcal{R}(s,a)b_t(s) ))
\end{equation}

A Bellman’s equation can be defined over the value function, opening the sum of Equation \ref{eq:pomdp_p1}. Thus, the optimal value function $V^*(b)$ can be obtained, as:

\begin{equation}
\label{eq:pomdp_p3}
    V^{*}(b) = \max_{a \epsilon \mathcal{A}}\left [ \sum_{s \epsilon \mathcal{S}}^{} \mathcal{R}(s,a)b_t(s) + \gamma \sum_{o \epsilon \Omega} Pr(o|b,\pi)V^{\pi}(b_{a}^{o} \right ].
\end{equation}

 Then, the optimal policy can be extracted according to:

\begin{equation}
\label{eq:pomdp_p4}
    \pi^{*}(b) = \arg \max_{a \epsilon \mathcal{A}}\left [ \sum_{s \epsilon \mathcal{S}}^{} \mathcal{R}(s,a)b_t(s) + \gamma \sum_{o \epsilon \Omega} Pr(o|b,\pi)V^{\pi}(b_{a}^{o} \right ].
\end{equation}

Assuming that the optimal policy is followed at every step after an action $a$ is taken in a belief state $b$: 
\begin{equation}
\label{eq:pomdp_p5}
    \mathcal{Q}^{\pi}(b,a) = \sum_{s \epsilon \mathcal{S}}^{} \mathcal{R}(s,a)b(s) + \gamma \sum_{o \epsilon \Omega} Pr(o|b,\pi)V^{\pi}(b_{a}^{o}), 
\end{equation}
\noindent where, the Q-function $\mathcal{Q}(a,b)$, represents the expected value where the expected immediate reward of action $a$ in a belief state $b$ is defined in the first term. The second term defines the sum of all posterior belief states, weighted by the probability of observing $o$ in those belief states \cite{Carmo2019}.

\textit{\textbf{Continuous state space}}

To extend POMDPs to continuous state space, the belief state update can be modeled as 

\begin{equation}
\label{eq:pomdp_cs}
b_{o}^{a}(s')=\frac{Pr(o|s',a) \int_{s \epsilon \mathcal{S}}(s'|s,a)b(s)ds}{ \int_{s' \epsilon \mathcal{S} } Pr(o|s',a)\int_{s \epsilon \mathcal{S}}Pr(s'|s,a)b(s) ds ds'}.
\end{equation}

The optimal value function $V^{*}(b)$ represents a continuous state space receiving observation space in the discrete state \cite{Carmo2019}.

\begin{equation}
\label{eq:pomdp_p6}
\begin{split}    
V^{*}(b) = \max_{a \epsilon \mathcal{A}}\left[\int_{s \epsilon \mathcal{S}}^{} \mathcal{R}(s,a)b(s)ds + \gamma \sum_{o \epsilon \Omega} Pr(o|b,\pi)V^{\pi}(b_{a}^{o})do \right].
\end{split}    
\end{equation}

\subsubsection{Methods for POMDPs}


POMDP is a mathematical framework that allows policy optimization and provides robustness to represent uncertainty within a system due to a lack of information, and specifically for non-deterministic and partially observable scenarios \cite{kurniawati2022partially}. Table \ref{tab:pomdp} shows several approaches adopted as solutions and their state spaces. 

There are several methods for evaluating information gain in POMDP path planning approaches, such as Expected Improvement \cite{Fernandez2022}, Shannon entropy \cite{Fernandez2022,Cliff2015}, weighted root mean square error \cite{Morere2017}, Kullback-Leibler (KL) divergence \cite{subramanian2019approximate,mao2020information}, or Rényi divergence \cite{nguyen2019trackerbots}.

\begin{center}
{\footnotesize
\begin{longtable}[H]{p{3.5cm}p{4cm}p{2.5cm}}

\caption{POMDP} \label{tab:pomdp} \\

\hline {\textbf{Reference}} & {\textbf{Approach}} & {\textbf{State Space}}\\ \hline 
\endfirsthead

    \citet{dallaire2009bayesian}     & Gaussian Process  & Continuous\\ \hline
   \citet{nguyen2019trackerbots}     & Monte Carlo sampling  & Discrete\\ \hline
    \citet{Fernandez2022}     & Gaussian Process and Monte Carlo planner & Continuous \\ \hline
    \citet{qiming2019modeling}     &  Monte Carlo sampling   & Discrete\\ \hline
   \citet{Hollinger2009}     &  Heuristic Search Value Iteration  & Discrete\\ \hline
    \citet{Morere2017}     & Bayesian Optimization and 
Monte-Carlo Tree Search (MCTS)  & Continuous \\ \hline
      \citet{Salhotra2021}     & Monte Carlo Tree Search  & Continuous\\ \hline
       \citet{Choudhury2020}     &  Monte Carlo Planning & Continuous\\ \hline
     \citet{reinhart2020learning}     &  Q-Learning &  Continuous\\ \hline
\end{longtable}}
\end{center}

\subsection{Gaussian Process Regression}
\label{sec:GP}

Gaussian Process \cite{RasWil2006} is a non-parametric Bayesian technique adopted to achieve non-linear regression from noisy observations. 
The Gaussian Process learns the transformation between input and output from observation. Further, it does not preserve an explicit model of the substantial phenomenon, which is why Gaussian has been identified as a non-parametric technique.  
It is Bayesian because it places \textit{prior} hypotheses on observed variables and updates them to produce a \textit{posterior} distribution as new information. Furthermore, it is considered a regression technique, as it produces continuous smooth outputs at arbitrary resolutions and indicates the best estimate and uncertainty inherent, given the current information~\cite{vivaldini2019uav}.

For the formulation problem, a dataset $\mathcal{D} = (X,\textbf{y}) = (\textbf{x}_i,y_i)_{i=1}^N$ composed of $N$ training inputs $\textbf{x}_i$ contains spatial coordinates and their probability $\textbf{y}_i$ for the observed environment. The correlation between the underlying function $f(.)$ and those probabilities are given by:

\begin{equation}
y(\textbf{x}_i) = f(\textbf{x}_i) + \epsilon,
\label{eq:undfunc}
\end{equation}

\noindent where $\epsilon$ is a Gaussian noise that describes how noisy the fit is to the actual observation. The Gaussian noise can be modeled as a zero-mean Gaussian distribution with variance $\sigma_n^2$, i.e., $\epsilon \sim \mathcal{N}(0,\sigma_n^2)$. 
Mean and covariance functions are also used to encode our prior knowledge concerning the underlying function. The mean function $m(\textbf{x}|\bm{\theta}_m)$ represents the space of possible functions fitting each point. The covariance function $\mu(\textbf{x}) = k(\textbf{x}_i,\textbf{x}_j|\bm{\theta}_k)$ models the correlation between any two given points~\cite{vivaldini2019uav}. The covariance function, together with the mean function, defines the Gaussian Process distribution.

\begin{equation}
    \label{eq:gp1}
    y \sim \mathcal{GP}(m(\textbf{x}|\bm{\theta}_m), k(\textbf{x}_i,\textbf{x}_j|\bm{\theta}_k))
\end{equation}

\noindent where $\bm{\theta}_m$ and $\bm{\theta}_k$ represent mean and covariance hyper-parameters, respectively.

In Gaussian Process Regression, the covariance function (kernel function) is the main building block for encoding structure into the model due to its influences on the Gaussian Process behavior. Different correlation techniques can be adopted for a specific scenario \cite{RasWil2006,duvenaud2013structure}, given that the choice greatly impacts the outcome, such as maximizing the likelihood or minimizing the prediction error. Among the most popular kernels used in Gaussian Process modeling are:

\textit{Squared Exponential Kernel Function}, also known as Radial Basis Function (RBF), focuses on producing nonlinear regression models for processes exhibiting continuous and smooth functional behavior \cite{Popovic2017online,samaniego2021bayesian}:

\begin{equation}
\label{eq:eqkf1}
k(\textbf{x}_i,\textbf{x}_j) =  \sigma_{m}^{2} \left(
\frac{ ( \textbf{x}_i - \textbf{x}_j )^T  L( \textbf{x}_i - \textbf{x}_j ) }{ 2  } \right).
\end{equation}

\textit{Rational Quadratic Kernel Function} results in a smooth prior on functions sampled from the Gaussian Process \cite{vivaldini2019uav,samaniego2021bayesian}:

\begin{equation}
\label{eq:rqkf}
k(\textbf{x}_i,\textbf{x}_j) = \left(
1 + \frac{ ( \textbf{x}_i - \textbf{x}_j )^T L( \textbf{x}_i - \textbf{x}_j ) }{ 2 \alpha } \right)^{-\alpha},
\end{equation}

\noindent where $\bm{\theta}_k=(\Sigma,\alpha)$, with $\Sigma$ and $\alpha$ being a length-scale diagonal matrix controlling smoothness, and $\bm{\theta}=\left(\sigma_m,\bm{\theta}_k,\sigma_n\right)$.

\textit{Periodic Exponential Kernel Function} results in random functions that are periodic almost everywhere (with the same period for every dimension):

\begin{equation}
\label{eq:eqkf2}
k(\textbf{x}_i,\textbf{x}_j) =  \sigma_{m}^{2} exp \left( - \frac{ 2 sin^2 (\pi T\sqrt{(\textbf{x}_i - \textbf{x}_j )^T L( \textbf{x}_i - \textbf{x}_j })  }{ \rho^2  } \right),
\end{equation}

\noindent where $L$ is a diagonal matrix.

Other kernels are also popular, such as Linear, Polynomial, Hilbert space \cite{francis2019occupancy}, and Matérn Class \cite{samaniego2021bayesian}, which are all dependent on a hyperparameter. More theoretical details can be found in \cite{RasWil2006}.

\subsection{Bayesian Optimization}
\label{sub:bo}

The two main components of Bayesian optimization are the search for the maximization information on an area (surrogate model) and information gain (acquisition function). The search maximization indicates the area with the highest incidences of targets. As the information gain acquires the average information about the area, we can say that it does so in the same proportion within the area. 
One of the essential aspects of the BO approach is the cost of choosing the next point based exclusively on the current state of the model. This can reduce uncertainty and allow for visitation to more promising areas in order to obtain information without considering the distance between the input points. Thus, more optimized solutions must consider not only the lowest cost, but also the solution which uses predictions about future information, as well as being able to use the objective function more effectively.

Bayesian Optimization is a sequential strategy used for finding the position $\textbf{x*}$ of the global maximum or minimum of an unknown function $f(.)$ (black-box function), represented by: 

\begin{equation}
    \label{eq:bo1}
    f(\textbf{x}): \mathbb{R}^D \rightarrow \mathbb{R}.
\end{equation}

\begin{equation}
    \label{eq:bo2}
    \textbf{x}^*: \arg \max_{x \epsilon \mathbb{R}^D} f (\textbf{x}).
\end{equation}

This function is expensive to evaluate and/or non-convex, meaning it can have multiple local maxima or minima ~\cite{Brochu2010}. It can be evaluated at any arbitrary point in its domain.

Using Bayes' theorem to combine obtained information and data with observation, BO produces new predictions of an underlying function $f (.)$ attempting to reach its maximum. 
Within each iteration, the next series of samples are chosen. These selections are done through an incomplete model carrying the data obtained in the previous interactions. The model is thereby improved by combining new samples while having no knowledge about its underlying function. 

The BO approach exploits all the evaluations it can get by building a surrogate model for modeling the objective function $f$, which represents the current belief about the unknown function. It uses this model to decide where to evaluate the function next in order to make the position of the maximum of the surrogate model as close as possible to the position of the maximum of the black-box function. The decision of the next position to be evaluated (sampled) in the domain $\mathbb{R}_D$ is performed by acquisition functions. It provides an equilibrium between exploration (where there is more uncertainty about the value of the function) and exploitation (where the current belief indicates the function has a high value).
Among the most essential choices to be made with BO is the kind of objective function to be used to approximate the black-box function and the acquisition function. This is done in order to balance exploration and exploitation by providing a measure for the utility of obtaining a new observation based on the surrogate model built using past observations~\cite{Khattab2019}. 

According to~\citet{samaniego2021bayesian}, a proper kernel and acquisition function selection make BO a suitable tool for providing confident models with a small number of samples. Some kernels are discussed further in Section \ref{sec:GP}. The acquisition function leads the search for the optimum in every iteration, reflecting the expected utility of sampling at each location in the domain. Consequently, higher values of the acquisition function should correspond to a greater probability of finding higher values of $f$. The most common acquisition functions are Probability of Improvement (PI), Information Gain, Expected Improvement (EI), and Upper Confidence Bound (UCB), which control a trade-off between exploration and exploitation while searching for the optimum. Among others, Entropy Search, Scaled Expected Improvement, Predictive Entropy Search, and the proposal for new acquisition functions (such as the Output-Informed Acquisition Functions to the exploration planning proposed by \cite{Blanchard2022}, as well as the Bayesian Optimization for path planning proposed by \cite{vivaldini2016,vivaldini2019uav}).

\subsection{Bayesian Optimization for Path Planning}
\label{sub:bo-path-planning}

\subsubsection{Acquisition function for path planning} 
The acquisition function focuses on minimizing any unvisited areas in order to produce a map of the environment accurately.  \citet{vivaldini2019uav} 
 also presented the coding of behavior through the acquisition function detailed below:
 
\begin{equation}
\label{eq:bopath}
h(\textbf{x}) = -\sigma_v^2 * \exp \left( -\frac{1}{2} \left( \frac{\mu - 0.5}{\sigma_l}\right)^2 \right).
\end{equation}

The Gaussian distribution employs a standard deviation $\sigma_l$, amplitude $\sigma_v$, and mean $0.5$. The negative sign is adopted to flip the Gaussian distribution, dealing with minimization uncertainties during the optimization process.

The traditional BO derivation is discrete, i.e. only the final destination of each iteration is considered. However, the authors extend it to a continuous domain \cite{Marchant2012} in which the trajectory between start and end points are also considered. This is of particular interest for the application in question because it enables the aircraft to obtain images as it navigates between points with no additional effort. A score $s$ for each trajectory $\mathcal{C}$ is calculated by the integration of the acquisition function over its length.

\begin{equation}
\label{eq:intcurve}
    s(\mathcal{C}(u,\bm{\beta}|h)) = \int_{\mathcal{C}(u,\bm{\beta})} h(u)du \text{,} 
\end{equation}

\noindent where $\bm{\beta}$ are trajectory coefficients and $u=[0,1]$. If Eq. (\ref{eq:intcurve}) cannot be analytically calculated \footnote{\cite{vivaldini2019uav} employed line segments as the template for trajectory calculations. However, Equation \ref{eq:intcurve} can be equally applied to any curve, such as to splines, as in \cite{Eger2001}.}
, approximations such as sampling or rectangle-rule quadrature \cite{Gau2012} may be used. Once the optimized trajectory $\bm{\beta}^*$ has been determined, samples $\{\textbf{x},y\}_{\mathcal{C}}$ are obtained throughout (i.e. at fixed-length intervals) and added to the GP model as new training points. The process is then repeated to produce a new optimized path based on the updated model. 


\section{Conclusions and Future Research Directions}
\label{sec:future}

Given promising advances in technology, we can assume that the future trend will have great potential for innovation and progress for autonomous robots to be able to handle not only the uncertainty of their systems but also overcome the challenges of dealing with uncertainty in the environment. Concerning an autonomous UAV, several major issues must be considered, including a large state, observation, action spaces, long planning horizon, and complex dynamics. Thus, this work contributes to the understanding of autonomous UAVs in the decision-making process for path planning in the Exploration and Informative Path Planning areas. 

First, we can highlight the definitions reported in this review to tackle exploration and informative path planning. As such, exploration is the collection of all information from the environment with the intention of performing the mapping. Informative planning is the collection of data from specific targets, although not necessarily from the entire environment. However, both collections also can produce area mapping. Both approaches for mapping can be adopted for further area monitoring, either for the general area or for a specific target. We can also understand that by monitoring an unknown area, we will obtain a mapping of what was observed. 

Many applications require solutions for Exploration and Informative Path Planning, and more and more autonomous robots are being adopted for these tasks. Decision-making in uncertain environments is related to how we acquire necessary information from the environment, and whether such representation is to be used for Informative or Exploration Path Planning. It is an easy task for human beings to evaluate and make decisions based on observations; however, many challenges are posed in this regard to the scientific research area of autonomous robotics. 

We observed the similarities in Informative and Exploration Path Planning of autonomous robot applications in uncertain environments, involving the same solution methods, albeit with modifications in the optimization, information gain, or objective functions. Additionally, modifications are presented in how to use information in the decision-making process to infer future approximations~\cite{folch2022,helan2022}, as well as in reducing the computational cost by adopting different sampling approaches in the objective function \cite{Fernandez2022}.

From a probability perspective, we have a probabilistic distribution on the variable amount of information that is being maximized. The proper way to assess this is to use information gain criteria and visit unexplored areas while re-observing the area for information accumulation. These methods seek to maximize information gain by looking at the accumulation of information and for which trajectory obtaining this accumulation might be possible. Additionally, they seek to minimize revisitation of areas already observed and obtain accuracy in their mapping.

Further, these methods are adopted not only to attempt to understand the area as a whole, but also to accumulate more information. They are also adopted for problems that involve searching for specific information from unknown regions; that is to say, observing areas where specific information is obtained and optimizing this information gain.

For autonomous robotics research, the Bayesian Optimization (BO) and Partially Observable Markov Decision Process (POMDP) help the decision-making process in both Exploration and Informative Path Planning problems. The BO approach reduces uncertainty in unexplored areas and can be used to solve both problems depending exclusively on the adopted acquisition function, given that the Bayesian inference updates the belief of a previous step with the new observations received.

The BO has two main components: information gain and search maximization over an area. Search maximization localizes the area with the highest incidence of targets. As the information gain component seeks to acquire the average information about an area, we can say that it does so in the same proportion throughout the area. 
One of the essential aspects of the BO approach is the cost of choosing the next point based exclusively on the current state of the model. This can reduce uncertainty and allow for visiting of more promising areas in order to obtain information without considering the distance between the input points. Thus, more optimized solutions must consider not only the lowest cost, but also the solution which uses predictions about future information, as well as being able to use the objective function more effectively.

The POMDP approach is a general methodology for the uncertainty decision process, where information is integrated into a sequential decision process through partial observations. Thereby, an action plan is obtained to achieve more promising information. It is, therefore, a decision-making problem in time intervals applied to environments in which the results of actions cannot be defined deterministically.
POMDPs are justified by their ability to examine environments where information is incomplete or imprecise, and the result of the POMDP depends on the quality of performance on the policy adopted \cite{golowich2022planning}. However, there may be failures in the observation model where the observation is uninformative and reports a low probability. Software tools for POMDP solving have been used in interfacing with typical robotics applications. Although some scalability issues in POMDPs have remained unchanged, studies show that existing methods may be capable of improving the scalability and robustness of POMDPs, thus highlighting the potential of POMDPs in robotics applications and presenting an endless line of potential research.

A new paradigm of adopting the similarities between Bayesian inference and path planning under uncertainty for the mapping of an unknown environment was introduced by \cite{farhi2022bayesian}. The inference (RUB Inference over the iSAM2 method) and the Belief Space Planning (BSP) were used in a POMDP configuration. The inference is efficiently updated using information from the previous planning stage in order to decide the subsequent action. This approach considers inconsistent data association on obtained and predicted measurements and future actions. Even when considering high-dimensional state spaces, the computation of inference time decreases run-time without affecting accuracy. Therefore, the unification between inference and decision-making under uncertainty presents new research directions, which can be applied to several branches of autonomous robotics.

Reinforcement Learning (RL) techniques have been used and suggested as methods for improving prediction in decision making \cite{Ruckin2022,farhi2022bayesian,golowich2022planning,shrestha2022spectrum}. In addition, a mathematical formulation of POMDP in the Exponential Time Hypothesis has been proposed previously \cite{golowich2022planning}.

 In applications with a large number of observation points, the clustering technique can be adopted for decision-making-based path planning. The clustering technique involves grouping data points (observations) into meaningful clusters, allowing for a more efficient exploration of the environment.
Then, clustering methods can be used in conjunction with this approach to reduce the complexity, making it easier to solve. A new research direction may explore how clustering techniques can be used to improve the efficiency of the decision-making-based path planning approach.

The Dual control for exploration and exploitation approach in \cite{Tan2022,chen2021dynamic,li2021concurrent,chen2022perspective} shares similarities with the Active Inference philosophy of neuroscience for perception interpretation, action, and learning. Therefore, future perspectives of research indicate a combination of both approaches. 

In our analysis, we can also verify the existing trend of dividing the solution into two layers: local planning and global planning \cite{petit2022tape,dang2020graph,cao2021tare,dang2019explore,cao2021exploring}, as well as division into a global exploration layer and a two-stage path planning layer \cite{zhou2021fuel,zhao2022faep}. Some approaches merge the strands by adopting an exploratory layer and an IPP to obtain better results \cite{Schmid2022}; others adopt both approaches, although with a focus on just one \cite{Blanchard2022}. This two-layer approach has the potential to improve the results of exploration and exploitation, and future directions of the research can focus on balancing exploration and exploitation between the layers. The two-layer approach has the potential to improve the results of exploration and exploitation, and future directions of the research can focus on balancing exploration and exploitation between the layers.

Future challenges for these approaches include the aspects of time-varying, non-periodic, spatiotemporal, and/or higher spatial variability in the decision process. In the time-varying aspect, the behavior of observations/information is time-dependent, so the planning will respond differently when acquiring the same observation as an input in $t+1$. In the non-periodic aspect, the observation is obtained randomly. In the spatiotemporal aspect, space and time data for obtaining an observation or performing an action influence the path planning, meaning there is a time dependence on the environment which can encode into spatiotemporal relationships. Regarding the spatial variability aspect, observations measured at different positions in space can show different information.
   
In the aforementioned approaches, several works have been adopted for Multi-UAV applications~\cite{manjanna2022scalable,Hardouin2020,li2022multi,Lei2022,Hayat2020,Ghassemi2022,Apostolidis2022,Cho2018,Wu2022,Xing2022,Bailon2022,sarkar2021novel,wang2022multi}. The theoretical analysis of robot coordination behavior for a better sampling of information gain is a challenging future direction of research for the scientific community, to understand the limitations of current approaches and develop new strategies to improve the efficiency of information gain. In \textit{Exploration or Informative Path Planning} applications, a UAV can be utilized to efficiently learn how to estimate the distribution and coordination of tasks from a certain previous observation. Future directions of research can focus on improving the accuracy of the estimation process and exploring new applications for this approach. In a decentralized architecture topic, the problem of maximizing information gain can be addressed where the prior knowledge of each UAV is unknown and the revisitation to explored areas needs to be minimized, taking into account the communication aspects in the gain information among robots. Future work on this topic can focus on developing efficient methods for estimating the distribution and coordination of tasks from a certain previous observation.

It can be concluded that, based on existing research, there are numerous promising research streams enabling the contribution and advancement of autonomous UAVs, where the decision-making for the level of the UAV's autonomy is a crucial factor.


\section*{Acknowledgment}

This work was supported by the CTU [grant number SGS23/177/OHK3/
3T/13]; the Czech Science Foundation (GAČR) [research project number 23-06162M]; the Ministry of Education of the Czech Republic through the OP VVV funded project "Research Center for Informatics" [grant number CZ.02.1.01/0.0/0.0/16 019/0000765]; the European Union’s Horizon 2020 research and Innovation programme AERIAL-CORE [grant number 871479].





\bibliographystyle{elsarticle-num-names} 
\bibliography{main}






\end{document}